\documentclass{bmvc2k}

\title{Guided Attention for Interpretable Motion Captioning}

\addauthor{Karim Radouane}{karimradouane39@gmail.com}{1}
\addauthor{Julien Lagarde}{julien.lagarde@umontpellier.fr}{2}
\addauthor{Sylvie Ranwez}{sylvie.ranwez@mines-ales.fr}{1}
\addauthor{Andon Tchechmedjiev}{andon.tchechmedjiev@mines-ales.fr}{1}

\addinstitution{EuroMov Digital Health in Motion, University of Montpellier, IMT Mines Ales, Ales, France}
\addinstitution{ EuroMov Digital Health in Motion, University of Montpellier, IMT Mines Ales, Montpellier, France}

\runninghead{RADOUANE et al.,}{Interpretable Motion
Captioning}

\usepackage{cleveref}

\usepackage{graphicx}
\usepackage{subcaption}
\usepackage{booktabs}
\usepackage{colortbl}

\usepackage{multicol}
\usepackage{multirow}

\usepackage{placeins}
\usepackage{amsfonts}
\usepackage{colortbl}

\begin{document}

\maketitle

\begin{abstract}
Diverse and extensive work has recently been conducted on text-conditioned human motion generation. However, progress in the reverse direction, motion captioning, has seen less comparable advancement. In this paper, we introduce a novel architecture design that enhances text generation quality by emphasizing interpretability through spatio-temporal and adaptive attention mechanisms. To encourage human-like reasoning, we propose methods for guiding attention during training, emphasizing relevant skeleton areas over time and distinguishing motion-related words. We discuss and quantify our model's interpretability using relevant histograms and density distributions. Furthermore, we leverage interpretability to derive fine-grained information about human motion, including action localization, body part identification, and the distinction of motion-related words. Finally, we discuss the transferability of our approaches to other tasks. Our experiments demonstrate that attention guidance leads to interpretable captioning while enhancing performance compared to higher parameter-count, non-interpretable state-of-the-art systems. The code is available at: \url{https://github.com/rd20karim/M2T-Interpretable}.

\end{abstract}

\section{Introduction}
\label{sec:intro}

Motion-to-language datasets such as KIT-ML \cite{Plappert2016} have garnered significant interest in motion-language applications. The motion captioning task is closely related to video captioning. However, human pose representation reduces the amount of data that needs to be processed and helps the model focus on the most important aspects of human motion, enabling more effective descriptions of human activities. In this context, the motion captioning task aims to generate natural language descriptions from sequences of human poses. Compared to the significant work done in vision-based captioning, which has seen different interpretable approaches identifying zones in images or videos that most contribute to the captions \cite{image_captioning_survey,xiao-etal-2019-guiding}, interpretability has been relatively less emphasized in motion captioning methods \cite{Guo_2022_TM2T,Plappert2017}. Nonetheless, an interpretable model holds significant importance in ensuring model reliability, offering explainable predictions for users, understanding model limitations. In this paper, taking inspiration from captioning approaches in vision, we devise a novel interpretable motion captioning system incorporating spatio-temporal and adaptive attention mechanisms. Moreover, the attention is guided to better match the human perception. To the best of our knowledge, this is the first interpretable system for motion captioning at both spatial and temporal levels. We demonstrate the performance of our interpretable captioning approach on available benchmarks: KIT Motion-Language Dataset \cite{Plappert2016} and HumanML3D \cite{Guo_2022_CVPR}, using common metrics, in alignment with current best practices for this task. Our contributions are summarized as follows:

\begin{itemize}
    \item We propose an interpretable architecture design that offers a transparent reasoning process, mimicking human-like attention perception and analysis, in contrast to  black box approaches.
    
    \item Novel formulation of adaptive gating mechanism, along with spatio-temporal attention in the context of human motion captioning.
    
    \item  We propose methodologies for adaptive and spatial attention supervision, aligned with our human skeleton partitioning method, which divides the body into six parts. This partitioning integrates separated local and global motion representations, aiming to enhance interpretability.
    
    \item We conduct extensive evaluations and analysis of our model's interpretability, involving qualitative assessments through attention maps and quantitative analyses utilizing specific proposed histograms and density distributions. Moreover, we demonstrate the capacity to leverage resulting model interpretability for action localization, body part identification, and distinguishing motion-words. 
   
\end{itemize}

\section{Related Work}
\label{sec:related}

\paragraph{Motion Captioning.}
 The first approach on the KIT-ML dataset \cite{Plappert2016} was introduced by \citep{Plappert2017} using a bidirectional LSTM. Later systems mainly focused on motion generation \citep{Lin2018, Ghosh2021, petrovich22temos}, but motion captioning has seen a resurgence with the introduction of HumanML3D \cite{Guo_2022_CVPR}. This dataset was firstly used for motion captioning by \citep{Guo_2022_TM2T}, which proposes learning motion tokens using VQ-VAE that are then mapped to word tokens through a Transformer \cite{Vaswani2017}. The results of this approach was not high specifically on KIT-ML (BLEU@4 =$18.4\%$). Then, \cite{Radouane_2023} slightly improved text generation results using a combination of Multilayer Perceptron (MLP) and Gated Recurrent Unit (GRU).
 Multitask learning was introduced in MotionGPT \cite{jiang2024motiongpt}, but the disparity in tasks prevents fair comparisons. However, this strategy negatively impacted motion captioning, yielding a low BLEU@4 score of 12.47\% on HumanML3D and no reported results on the KIT-ML dataset.

\paragraph{Adaptive attention.} Attending to the input (\textit{e.g.,} image) for the generation of non-visual words can be misleading and degrading to the performance of attention networks. To alleviate this problem, \cite{Jiasen2017} propose a formulation for a learnable gate variable $\beta$. The variable $\beta$ is learned to choose either to rely on the image features or only on the context of language generation through the visual sentinel vector. For motion captioning, this is particularly relevant as only specific words ("walks", "throw", etc.) need to access motion input information during prediction time, in contrast to non-motion words ("a", "the", etc.).

\paragraph{Guided attention.} Attention mechanisms can focus on incorrect areas of the input or on regions with a strong bias that aren't particularly meaningful for human interpretation. To mitigate these limitations \cite{Liu2016} propose attention supervision, a technique aimed at improving the performance and accuracy of image captioning models. This approach leads to more relevant attention maps, thereby enhancing interpretability. In the context of video captioning, spatial guiding of attention has also been shown to improve captioning performance \cite{Yu2017}.

\section{Methods}
\label{sec:method}

We first present the general model architecture for our captioning approach (\Cref{sec:model_archi}), followed by more in-depth presentations of our formulations for spatial and adaptive attention, as well as our attention guidance methodology (\Cref{sec:spat_adapt_superv}).

\subsection{Architecture design for motion captioning}
\label{sec:model_archi}

\begin{figure*}[t]
    \centering
    \includegraphics[width =.98\textwidth]{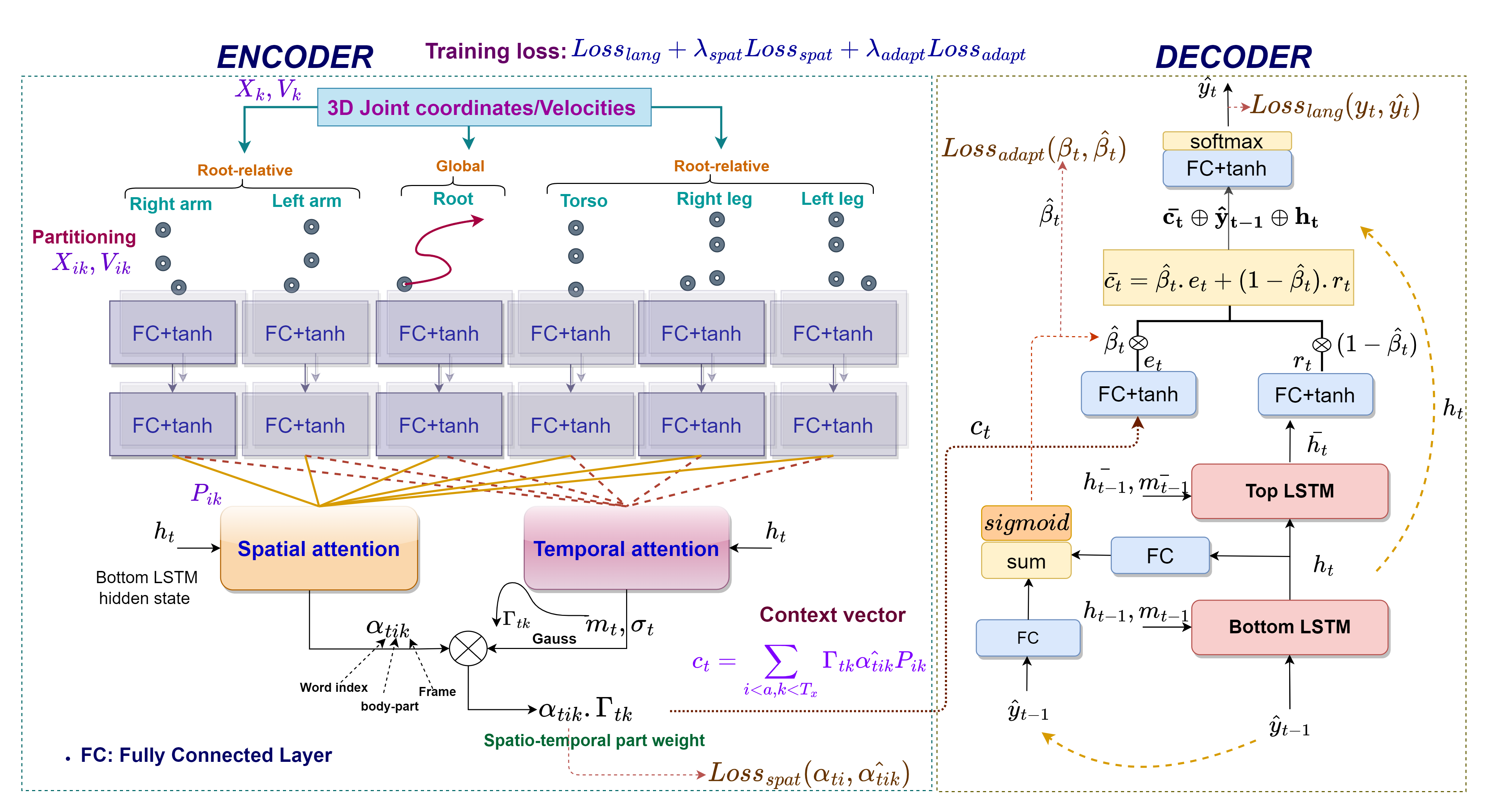}
    \caption{The encoder branch encodes frame-wise part-based motion representations from joint positions (\(X_{ik}\)) and velocities (\(V_{ik}\)), while the decoder branch takes as input (previous token \(\hat{y_{t-1}}\), previous state (\(h_{t-1}\), \(m_{t-1}\))) and estimates the relative importance (\(\hat{\beta_t}\) gate) of motion information to consider for word prediction $\hat{y_t}$. Spatial (\(\hat{\alpha_{tik}}\))  and temporal attention $(\Gamma_{tk})$ are computed from encoded part embeddings  $P_{ik}$ and \(h_{t}\). The spatio-temporal weights are used to compute the context vector \(c_t\) which is then passed to the decoder adaptive gate.  \(Loss_{lang}\) the cross entropy between predicted, and target words is the main loss. We propose to guide spatial and adaptive attention with \(Loss_{spat}\) and  \(Loss_{adapt}\).
}
    \label{fig:archi_sptemp}
\end{figure*}

Our model, summarized in \Cref{fig:archi_sptemp}, is composed of an encoder block, a spatio-temporal attention block and a text generation/decoder block incorporating an adaptive attention mechanism. 

Let $\textbf{X} \in \mathbb{R}^{T_x \times J \times D}$ be the input sequence of motion features of $T_x$ time steps, where $J$ is the number of joints in the skeleton and $D$ is the number of spatial dimensions. We note by $X_k$ the 3D joints positions and $V_k$ their corresponding velocities at frame time $k$. 

\textbf{Skeleton partitioning.} We group the joints in 6 body-parts: Left Arm, Right Arm, Torso, Left Leg, Right Leg, Root. We convert the global coordinates to root-relative coordinates, except for the root itself, which describes the global trajectory of the motion. $X_{ik}$ denotes the group of joints of part $i$ for every frame $k$ as described in \Cref{fig:archi_sptemp}.

\textbf{Encoder.} Each of the six body parts is embedded by two linear layers followed by  $tanh$ activations, as illustrated in \Cref{fig:archi_sptemp}. Each linear layer (FC) encode positions $X_{ik}$ and velocities $V_{ik}$ separately. The final embedding $P_{ik}$ for a given part $i$ and frame $k$ is the concatenation of the position and velocity embeddings. We note by $P$ the frame-level motion features of all human body parts. $P  \in \mathbb{R}^{ T_x \times a \times h_{enc} }$ where $h_{enc}$ the dimension of the final output encoder and $a=6$ is the number of body parts $(P = Enc ( X ))$. 

\textbf{Decoder.} We adopt a two-LSTM decoder configuration, a \textit{Bottom LSTM} for learning attention weights and language context and a \textit{Top LSTM} for final word generation based on the relevant information extracted from language and motion. We note by  $\mathbf{y} = (y_1, \ldots, y_{T_y}),\mbox{ }y_i \in \mathbb{R}^{K_y}$ the sequence of words describing the motion.  Let $h_t \in \mathbb{R}^{h_{dec}} $ be the decoder hidden state of the bottom LSTM for a word $w_t$ in the sequence and $\bar{h_t}$ for the Top LSTM. We note by $K_y$ the size of the target vocabulary. $T_x$ and $T_y$ are respectively the length of the motion sequence and the length of its description.
The decoder $Dec$ is used to predict the next word $y_{t}$ given the adaptive context vector described by $\bar{c_t}$ and the previous word $y_{t-1}$ and the bottom hidden state $h_t$.

\begin{equation}
      \label{eq:decoder_prob}
     p(y_t \mid \left\{ y_1, \cdots, y_{t-1} \right\}, \bar{c_t}) = Dec(y_{t-1}, h_{t}, \bar{c_t}) 
\end{equation}

The context vector $c_t$ is computed by a spatial-temporal attention mechanism, where temporal attention determines \textbf{when} to focus attention, and spatial attention determines \textbf{where} to focus in the body part graph. In the following, we note by $P^{*} \in { \mathbb{R}^{ h_{enc} \times a \times T_x }}$ the permutation of $P  \in \mathbb{R}^{ T_x \times a \times h_{enc} }$.

\textbf{Temporal attention.} The temporal weights are computed from extracted motion features $P^{*}$ and the current decoder hidden state $h_t$.
\begin{align}
    \boldsymbol{z}_t & =\boldsymbol{w}_h^T \tanh (\boldsymbol{W}_p \boldsymbol{P^{*}}+ep(\boldsymbol{W}_h \boldsymbol{h}_t))   \\ 
    \boldsymbol{\gamma}_t & =\operatorname{softmax}(\boldsymbol{z}_t)    
\end{align}

Here $\boldsymbol{W}_p \in \mathbb{R}^{d \times h_{enc}}$,$\boldsymbol{W}_h \in  \mathbb{R}^{d \times h_{dec}}$ and $\boldsymbol{w}_h \in \mathbb{R}^{d \times 1}$ are learnable parameter, $ep$ is an expansion operator mapping to $d \times a \times T_x $, and $a$ the number of body parts. Moreover, $\gamma_t$ is the temporal attention weights for the word generated at time $t$. With the above formulation, we often have discontinuities in the attention maps, yet such discontinuities are undesired, as the action happens continuously in a given frame range. The distribution of attention weights for a \textit{particular motion word} can be modelled as a Gaussian distribution with a learnable mean and standard deviation. The mean $m_{t}$ and standard deviation $\sigma_{t}$ are computed from the previous temporal attention weights $\gamma_{tk}$, which are replaced by $\Gamma_{tk}$ during training in this case (See \Cref{fig:archi_sptemp}). Intuitively, the mean $m_t$ will approximately represent the center time of action duration described by a motion word $w_t$, and the spread of the distribution approximately corresponds to the duration of the action.

\begin{equation}
    \Gamma_{tk} =\exp{(-\frac{(k-m_t)^2}{2{\sigma_t}^2})}
\end{equation}

\textbf{Spatial attention.}
Spatial weights are computed for each body part (Torso, left/right arm, left/right leg) as follows:
\begin{align}
    \boldsymbol{s}_t & =\boldsymbol{{w_{s}^T}} \tanh \left(\boldsymbol{W_{{p}_s}} \boldsymbol{P^{*}}+ep(\boldsymbol{W_{{h}_s}} \boldsymbol{h}_t) \right) \\ 
    \boldsymbol{\alpha}_t & =\operatorname{softmax}\left(\boldsymbol{s}_t\right)
\end{align}

Here $s_t \in \mathbb{R}^a$. The learnable parameters are $\boldsymbol{W}_{p_s} \in \mathbb{R}^{d\times h_{enc} }$,$\boldsymbol{W}_{h_s} \in  \mathbb{R}^{d \times h_{dec}}$  and $\boldsymbol{w}_s \in \mathbb{R}^{d \times 1}$.  We note by $\alpha_{t,m,k}$ the spatial attention score for part $m$ of the skeleton graph at frame $k$ for the word generated at time $t$. Thus, explicitly  $\alpha_t = [\alpha_{t,1,1}, \alpha_{t,1,2}, \cdots, \alpha_{t,a,T_x}]$.

\textbf{Adaptive attention.} Non-motion words, particularly grammatical words, do not carry any information about the movement. Consequently, we propose to learn a gating variable $\hat{\beta_t} $ to decide the proportion to which to use language context over motion features.
\begin{equation}
    \hat{\beta_t}  = sigmoid(W_b^h.h_{t}+W_e.(E\hat{y}_{t-1}))
    \label{eq:adaptive_beta}
\end{equation}

Where $W^h_b \in \mathbb{R}^{1\times h_{dec}}, W_e \in \mathbb{R}^{1\times d_{emb}}$ are learnable matrices. $E \in \mathbb{R}^{d_{emb} \times K_y} $ refers to embedding matrix of target words. The gating variable depends on the hidden state, which encodes residual information about generated words up to the time step $t$, as well as on the embedding of the previous word, as detailed in \Cref{eq:adaptive_beta}.

\textbf{Context vector.} The context vector is derived by weighting the motion features with spatial and temporal attention weights, and averaging across the frame-time dimension \ref{eq:cteq1}.

\begin{equation}
    c_t=\sum_{k=1}^{T_x} \sum_{i=1}^{a}\Gamma_{tk}\alpha_{tik} P_{ik}
    \label{eq:cteq1}
\end{equation}

The motion $c_t$ and language information $\bar{h_t}$ are embedded into the same space through an linear layer with $tanh$ activation (for bounded values in [-1,1]), giving $\bf e_t$ and $\bf r_t$ respectively.

\textbf{Adaptive context vector.} Given by \Cref{eq:cteq2}. When $\hat{\beta_t}=1$ the model uses full motion information and when $\hat{\beta_t}$ is close to $0$ the model relies more on language structure. 
\begin{equation}
    \bar{c_t} = \hat{\beta_t} .e_t + (1-\hat{\beta_t}).{r_t} \label{eq:cteq2}
\end{equation}

Finally, the probability outputs are computed  as in \Cref{eq:p_yt_ct}, similarly to previous work on video captioning \cite{Song2018}, except we include the bottom hidden state. This ensures that the language information of previously generated words is always present, which is important for correct syntax, even for motion words (e.g. jogs, jogging…).

\begin{align} \label{eq:p_yt_ct}
    & p(\hat{y}_t \mid \hat{y}_{1:t-1}, \hat{c_t}) = softmax(\tanh{(W_f.concat([\hat{c_t};\hat{y}_{t-1};h_t]))}) 
\end{align}

\subsection{Spatial and adaptive attention supervision}
\label{sec:spat_adapt_superv}
To our knowledge, simultaneous supervision of attention mechanisms with an adaptive gate and spatial attention has never been applied to captioning tasks, particularly motion captioning. Below, we provide a formal definition of how the losses for attention supervision are formulated.

\textbf{Language loss.} The standard loss for motion-to-text generation is defined as the cross entropy between the target and predicted words:

\begin{equation}
    Loss_{lang} = -\sum_{t=0}^{T_y-1} y_{t} \cdot \log(\hat{y}_{t})
    \label{eq:loss_lang}
\end{equation}

\textbf{Adaptive attention loss.} 
To build a ground truth for adaptive attention, we define mapping rules to distinguish between motion words, action verbs and qualifying adjectives (\textit{e.g.,} walk, circle, slowly) from non-motion words (\textit{e.g.,} of, person). We assign $\mathbf{\beta_t=1}$ for motion words and $\mathbf{\beta_t=0}$ for non-motion words (See Supp.\ref{supp:gt_generation}).

\begin{equation}
    Loss_{adapt} =  -\sum_{t=0}^{T_y-1} \beta_t \log(\hat{\beta_t})+(1-\beta_t)\log(1-\hat{\beta_t}) 
\end{equation}

\textbf{Spatial attention loss.} The predicted attention score is $\hat{\alpha_{tik}}$ for a given word $w_t$ and part $i$ of the source motion at the frame $k$. The loss is formulated in \Cref{eq:loss_sp}, where $N_{y}$ is a normalization factor that count the number of supervised words for a given target description $y$ (See Supp.\ref{supp:gt_generation} for attention guidance strategy).

\begin{align}\label{eq:loss_sp}
    Loss_{spat} = - \frac{1}{N_y} & \sum_{i,t,k} {\alpha_{ti}} \log(\hat{\alpha_{ti}}) + (1-{\alpha_{tik}}) \log(1-\hat{\alpha_{tik}})  
\end{align}

\textbf{Global loss.} To define the global loss, we add the loss terms for spatial attention \(loss_{spat}\), adaptive attention gate \(loss_{adapt}\) guidance, respectively weighted by $\lambda_{spat},\lambda_{adapt}$, to control their contributions.

\begin{equation}
    Loss = loss_{lang}+\lambda_{spat}.loss_{spat}+ \lambda_{adapt}.loss_{adapt} 
    \label{eq:global_loss}
\end{equation}

\section{Experiments}
We consider the commonly used benchmarks KIT-ML \cite{Plappert2016} and the HumanML3D (HML3D) \citep{Guo_2022_CVPR} (Dataset statistics in Supp.\ref{supp:dataset}). We conduct ablation studies on both datasets to determine the impact of adaptive and guided attention, followed by a detailed analysis of our model's interpretability.

\textbf{Ablation Study.} We configure a search space for $(\lambda_{spat},\lambda_{adapt})$ and run the search using WandB \citep{wandb}. \Cref{tab: bleu_results} quantifies the impact of attention guidance. Due to space constraints, more results can be found in Supp.\ref{supp:hyperparam}, and additional detailed analysis regarding the effectiveness of our architecture components can be found in Supp.\ref{supp:abla_archi}).

\textbf{Hyperparameters.}
\label{sec:hyperparameters}
For both KIT-ML and HumanML3D datasets, we set respectively the word embedding size and decoder hidden size  to (\(d_{emb} = 64, h_{dec}=128\)) and ($d_{emb} = 128$, $h_{dec}=256$), respectively. Additionally, the output dimension of each fully connected layer \(FC_i\) is \(128\) for layer 1 and \(64\) for layer 2 in KIT-ML, and \(256\) for layer 1 and \(128\) for layer 2 in HumanML3D. After concatenation, we obtain \(128\) and \(256\) joint-velocity features per frame  respectively for KIT-ML and HML3D.

\begin{table}[t]\centering
\resizebox{\linewidth}{!}{
\begin{tabular}{lccccccc}
\toprule
\bf Dataset &  $\lambda_{\mbox{\bf spat}}$ & \bf $\lambda_{\mbox{\bf adapt}}$ &  \bf BLEU@1 & \bf BLEU@4 & \bf CIDEr & \bf ROUGE-L & \bf BERTScore \\ \midrule
 \multirow{3}{*}{\bf KIT-ML}    & 0 & 0 & 57.3 & 23.6 & 109.9 & 57.8 & 41.1 \\
                                & 0 & 3 & 56.3 & 22.5 & 108.4 & 56.5 & 39.8 \\
                                & \bf 2 & \bf 3 & \bf 58.4 & \bf 24.4 & \bf 112.1 & \bf 58.3 & \bf 41.2 \\ \midrule
\bf \multirow{3}{*}{\bf HML3D}    & 0 & 0 & 69.3 & 24.0 & 58.8 & 54.8 & 38.7 \\
                                & \bf 0 & \bf 3 & \textbf{69.9} & \textbf{25.0} & 61.6 & \textbf{55.3} & \bf 40.3 \\
                                & 2 & 3 & 69.2 & 24.4 & \textbf{61.7} & 55.0 & \textbf{40.3} \\
 
\bottomrule

\end{tabular}
 }

\caption{Results for different supervision modes, where $\lambda_{spat}=\lambda_{adapt}=0$ represents the case without any attention guidance  for comparison. The gate (adapt) and spatial (spat) supervision, perform well when used together on KIT-ML (small). For HumanML3D adaptive attention was always beneficial, but guided spatial attention slightly degraded exact matching scores (BLEU@4, ROUGE) compared to  only adaptive attention  (Detailed experimented values in Supp.\ref{supp:hyperparam}). The impact is more significant on the interpretability aspect \Cref{sec:intepre_analysis}. }
\label{tab: bleu_results}
\end{table}

\begin{table*}[t] 
\resizebox{\linewidth}{!}{

\centering

{
    \begin{tabular}{cccccccc}
    \toprule
    \bf Dataset  & \bf Model  & \bf BLEU@1      &  \bf BLEU@4 & \bf ROUGE-L & \bf CIDEr & \bf BERTScore\\ \midrule
    \multirow{5}{*}{\bf KIT-ML} 
                            ~   
                        ~       & SeqGAN \cite{Goutsu2021}         &  3.12      & 5.20 & 32.4 & 29.5    & 2.20 \\ 
                        ~       & TM2T \cite{Guo_2022_TM2T}    &  46.7      & 18.4 & 44.2 & 79.5     & 23.0 \\
         ~       & MLP+GRU \cite{Radouane_2023}      & 56.8 & \textbf{25.4} & \textbf{58.8}  & \textbf{125.7} & \textbf{42.1} \\ \cline{2-7}
                       ~  & \textbf{Ours-[spat+adapt](2,3)}     & \textbf{58.4} & 24.7  & 57.8 & 106.2 & 41.3 \\
                             ~ & \textbf{*Ours-[spat+adapt](2,3)}       & \textbf{58.4} &  \underline{24.4} & 58.3  & \underline{112.1} & \underline{41.2}\\    
    \midrule
    \multirow{5}{*}{\bf HML3D}
                        ~       & SeqGAN \cite{Goutsu2021}          &  47.8      & 13.5 & 39.2 & 50.2   & 23.4 \\ 
                        ~       & TM2T \cite{Guo_2022_TM2T}   &  61.7      & 22.3 & 49.2 & \textbf{72.5}   & 37.8 \\  
           ~        & MLP+GRU\cite{Radouane_2023}     &  67.0 & 23.4 & 53.8  &  53.7 & 37.2 \\ \cline{2-7}

          ~         & \textbf{Ours-[adapt](0,3)}      & \underline{67.9} & \textbf{25.5}  & \underline{54.7} & \underline{64.6} & \textbf{43.2}  \\
                              ~           & \textbf{*Ours-[adapt](0,3)}         & \textbf{69.9} &  \underline{25.0} & \textbf{55.3}  & 61.6 & \underline{40.3}  \\
          \bottomrule
    
    \end{tabular}
}}

\caption{Text generation performance, assessed with beam size $2$ as in \cite{Guo_2022_TM2T}, while * indicate a greedy search. Our model performs better than Transformer-based (TM2T) method on both datasets and on HumanML3D compared to MLP+GRU.}

\label{table:comp_sota}
\end{table*}

\subsection{Evaluation and discussion}
\label{sec:eval_discuss}

Table \ref{table:comp_sota} presents the comparison to SOTA systems. Our approach performs significantly better than other state-of-the-art approaches without beam search on HML3D, including the Transformer TM2T. For KIT-ML dataset, MLP+GRU is slightly better than our approach in terms of NLP metrics. However, in terms of interpretability, our approach provides more information on the body parts involved in an action compared to MLP+GRU, which lacks spatial and adaptive attention. Therefore, in their case, the motion representation doesn't consider the skeleton graph structure and is always utilized for generating non-motion words that don't require motion information, which may lead to biased learning.

\subsection{Interpretability analysis}
\label{sec:intepre_analysis}
\begin{figure}[h]
    \centering
    \begin{minipage}[b]{0.45\textwidth}
        \centering
        \includegraphics[width=\columnwidth,height=3.5cm]{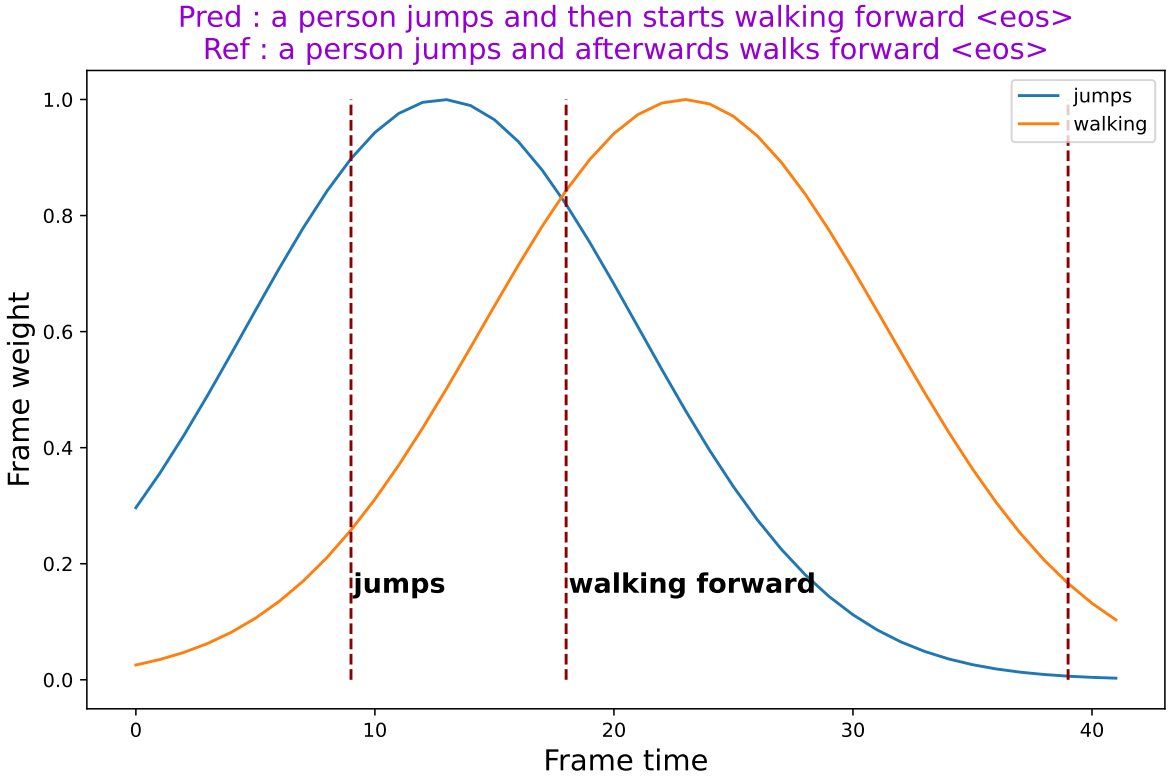}
        \caption{Temporal gaussian window displayed for different motion words given a prediction on KIT-ML.}
        \label{fig:TAG_figs}
    \end{minipage}
    \hfill
    \begin{minipage}[b]{0.5\textwidth}
    \includegraphics[width=\columnwidth,height=3.5cm]{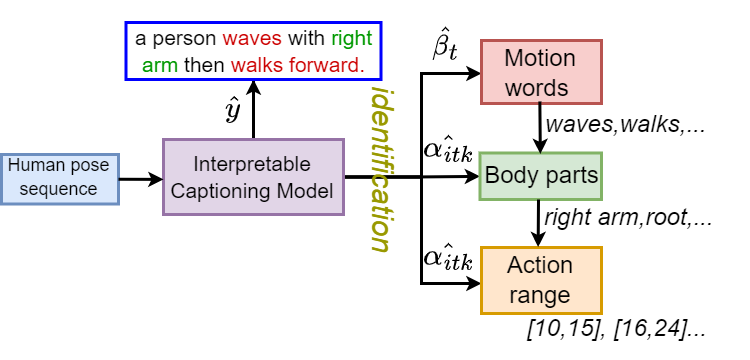}
    \caption{Interpretability use towards fine-grained captioning, based on spatial, temporal and adaptive attention scores.}
    \label{fig:interp_bloc}
    \end{minipage}
\end{figure}

In our context, interpretability is measured by the ability to establish a correspondence between learned attention mechanisms and human attention perception. In this section, we discuss the interpretability of learned attentions and how we can leverage interpretabiliy as illustrated in \Cref{fig:interp_bloc}. To demonstrate the role of each of the context vectors ${c_t}$ and LSTMs hidden states ($\bar{h}_t$,$h_t$), we fix the $\hat{\beta}$ value at $1$ and show a representative examples compared to adaptive gate in Table \ref{tab:comp_diff_beta}. Further analysis in Supp.\ref{supp:abla_archi}.

\begin{figure}[ht]
    \centering
    \begin{subfigure}{0.48\columnwidth}
        \centering
        \includegraphics[width=\columnwidth,height=4cm]{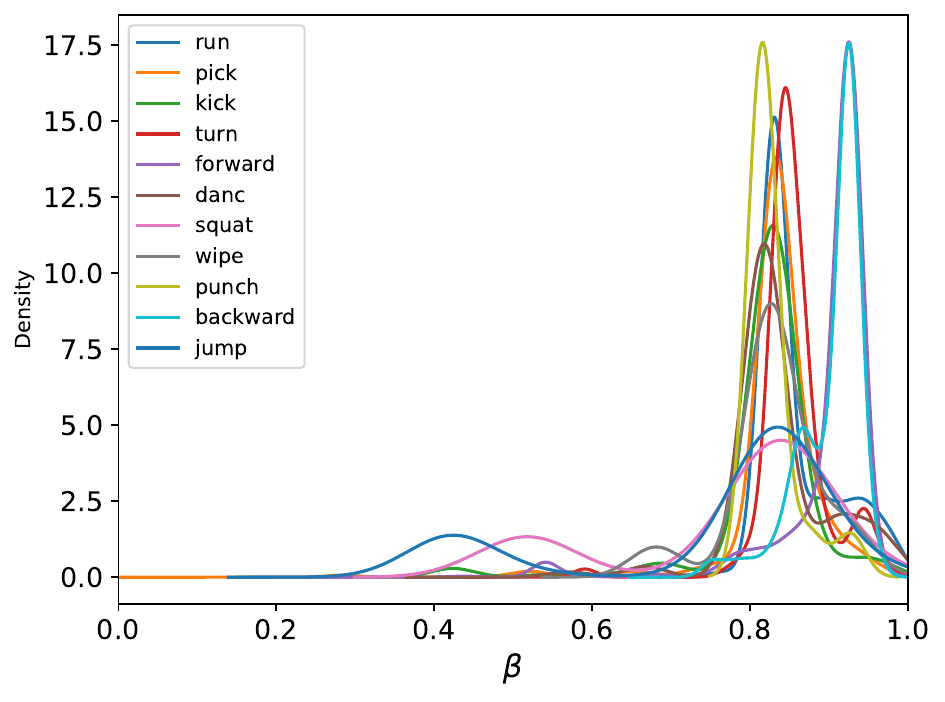}
        \caption{With gate supervision, motion information is correctly used frequently for motion-words generation.}
        \label{fig:w_gate_h3D_motion}
    \end{subfigure}\hfill
    \begin{subfigure}{0.48\columnwidth}
        \centering
        \includegraphics[width=\columnwidth,height=4cm]{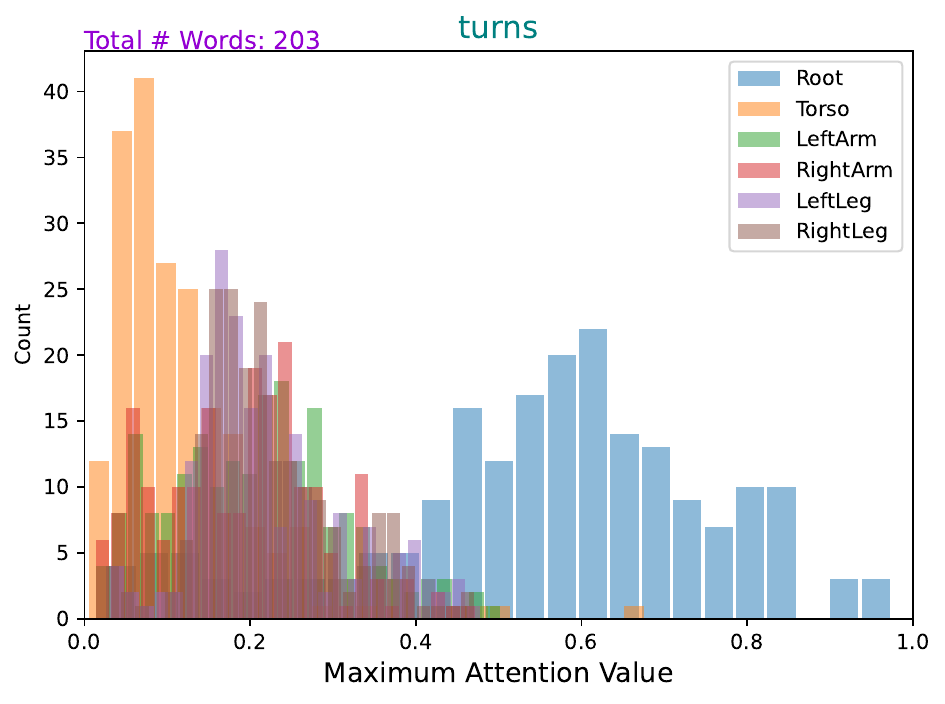}
        \caption{Attention is frequently focused on relevant parts: e.g. on Root (global trajectory) for word "turns".}
        \label{fig:w_spat_h3D_turn}
    \end{subfigure}
    \caption{$\hat{\beta}$ test set density distribution for a few motion words stems on HumanML3D and the temporal maximum body-parts attention histogram for word "turn".}
    \label{fig:w_vs_wo_gate_h3D_motion}
\end{figure}

\begin{table}[h]
\centering
\begin{tabular}{p{3.5cm}|p{3.5cm}|p{3.5cm}} 
\hline
$\hat{\beta}=1$ &\textbf{Adaptive $\hat{\beta}$} & \textbf{Reference} \\ \hline

\textcolor{blue}{walks forward} and \textcolor{teal}{sits down} \verb|<eos>| &
a person \textcolor{blue}{walks forward} \textcolor{olive}{turns around} and \textcolor{teal}{sits down} and \textcolor{olive}{gets back up} and \textcolor{olive}{walk back} \verb|<eos>| &
man \textcolor{blue}{walks forwards} stops \textcolor{olive}{turns around} and \textcolor{teal}{sits} then \textcolor{olive}{gets up} and \textcolor{olive}{walks back} \verb|<eos>| \\ \hline

\textcolor{blue}{jumping} up and down in place \verb|<eos>| & a person \textcolor{blue}{jumps} up and down \textcolor{olive}{multiple times} \verb|<eos>| & someone \textcolor{blue}{jumps} \textcolor{olive}{twice} and looks down at the ground \verb|<eos>| \\ \hline

punching \textcolor{blue}{boxing} and \textcolor{teal}{moving hands} around \verb|<eos>| & a person is \textcolor{blue}{boxing} with \textcolor{teal}{both hands} \verb|<eos>|  & a person standing up is making \textcolor{blue}{boxing} motions with their \textcolor{teal}{left and right arms} \verb|<eos>|

\end{tabular}
\caption{Comparison of the prediction when setting $\hat{\beta}=1$ and adaptive on HML3D-$(0,3)$ using human motion samples involving different actions.}
\label{tab:comp_diff_beta}
\end{table}

\begin{figure}[h]
    \centering
    \begin{subfigure}{.48\columnwidth}
        \centering
        \includegraphics[width=\columnwidth,height=4cm]{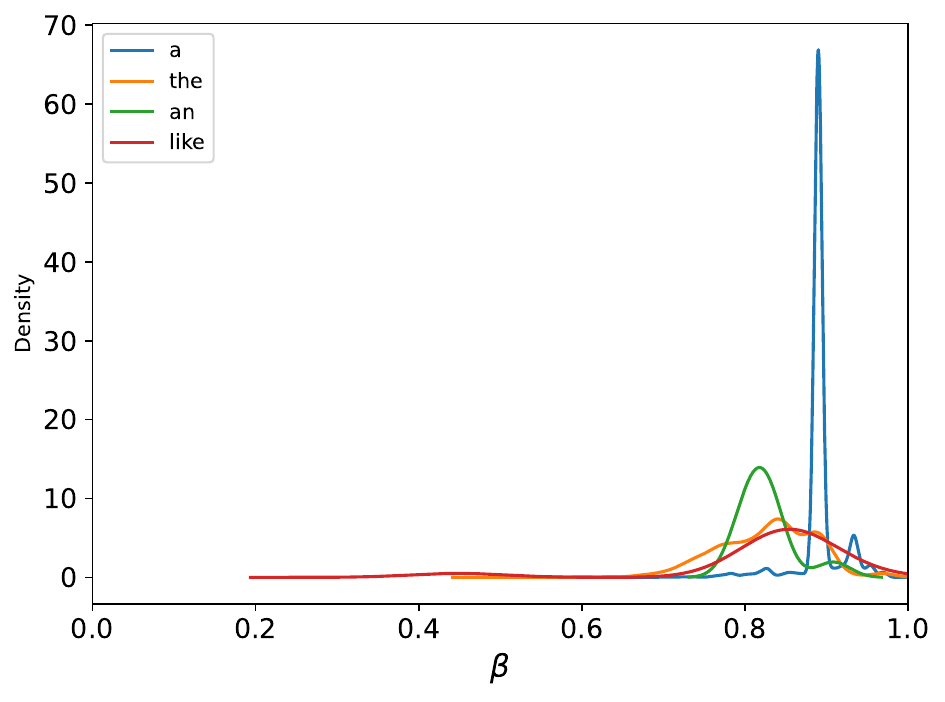}
        \caption{Without gate supervision, decoder uses frequently motion information even for non-motion words ($\beta$ frequently high).}
        \label{fig:w/o_gate_h3D}
    \end{subfigure} \hfill
    \begin{subfigure}{0.48\columnwidth}
        \centering
        \includegraphics[width=\columnwidth,height=4cm]{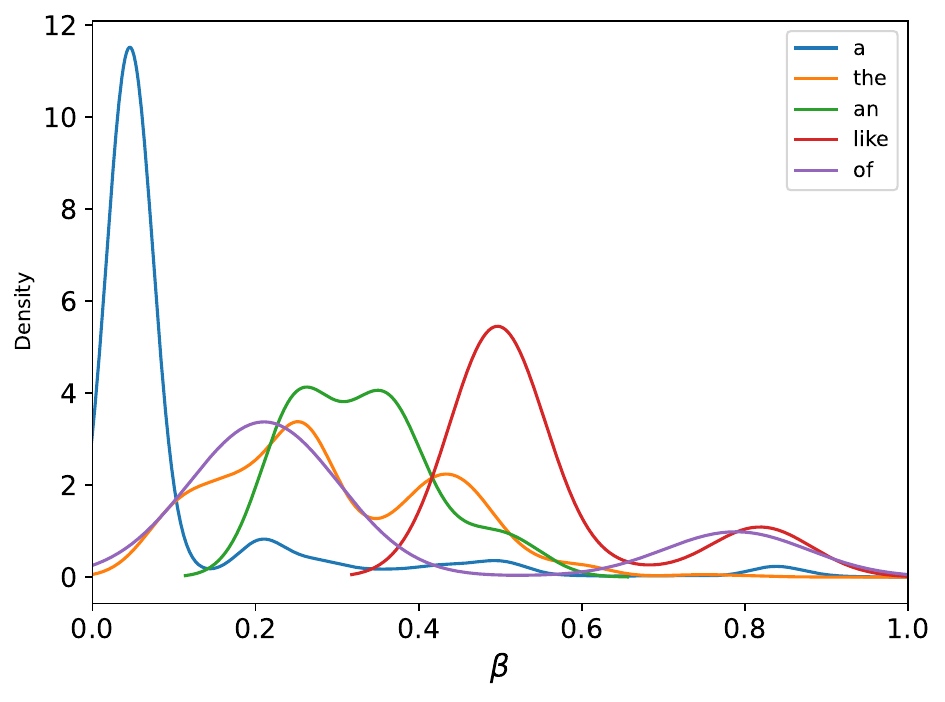}
        \caption{With gate supervision, the decoder uses correctly more language context for non-motion words ($\beta$ frequently small).}
        \label{fig:w_gate_h3D_nonmotion}
    \end{subfigure}
    \caption{$\hat{\beta}$ density distribution over test set for some non-motion words (stemmed) on HumanML3D.}
    \label{fig:w_vs_wo_gate_h3D_nonmotion}
\end{figure}

\begin{figure}[h]
    \centering
    \begin{subfigure}{0.46\columnwidth}
        \centering
        \includegraphics[width=\columnwidth]{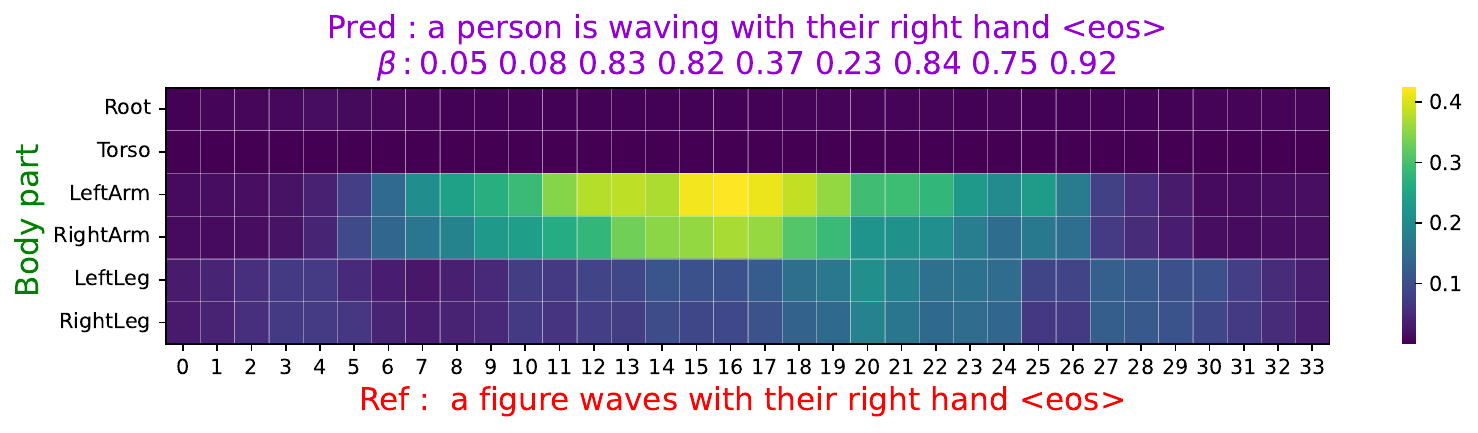}
        \caption{HML3D-(2,3), word \textit{waving} in range $[4,27]$. }
        \label{fig:spat}
    \end{subfigure} 
    \hfill
    \begin{subfigure}{0.49\columnwidth}
        \centering
        \includegraphics[width=\columnwidth]{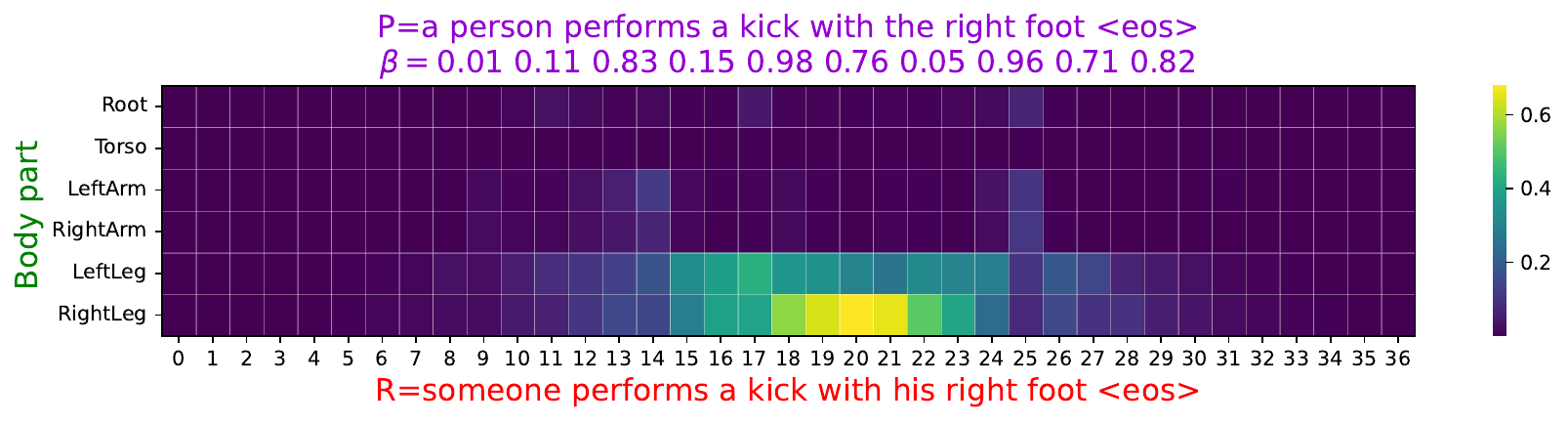}
        \caption{KIT-(2,3), word \textit{kick} in range $[16,26]$}.
        \label{fig:kick_h3d}
    \end{subfigure}
    \caption{Spatio-temporal attention maps for some words, with the color scale indicating attention score intensity per frame per body part. The model focalize correctly on relevant parts ((a).arms, (b).legs) at precise action timing and $\hat{\beta}$ values are semantically variable depending on the nature of predicted words as illustrated by the predictions in figures title (other examples in Supp.\ref{supp:visualizations}}
\label{fig:spat_temp}
\end{figure}

\textbf{Spatial / Adaptive attention impact.}
When training a model without guiding adaptive attention, we observe that $\hat{\beta}$ gate values frequently takes higher values for non-motion words (a:$0.9$, the:$0.8$) as illustrated in \Cref{fig:w/o_gate_h3D}. This behavior degrades performance, as seen in \Cref{tab: bleu_results} for both datasets. However, when we introduce adaptive gate supervision (cf. \Cref{fig:w_gate_h3D_nonmotion}), the model more frequently assigns less weight $\hat{\beta}$ to non-motion words and begins to learn how to make decisions automatically when to use the context vector, as illustrated also in Figure title \ref{fig:kick_h3d}, while guided spatial attention enhances the learned attention maps.

\begin{figure}[h]
    \centering
    \begin{subfigure}{0.48\columnwidth}
        \centering
        \includegraphics[width=\columnwidth,height=4cm]{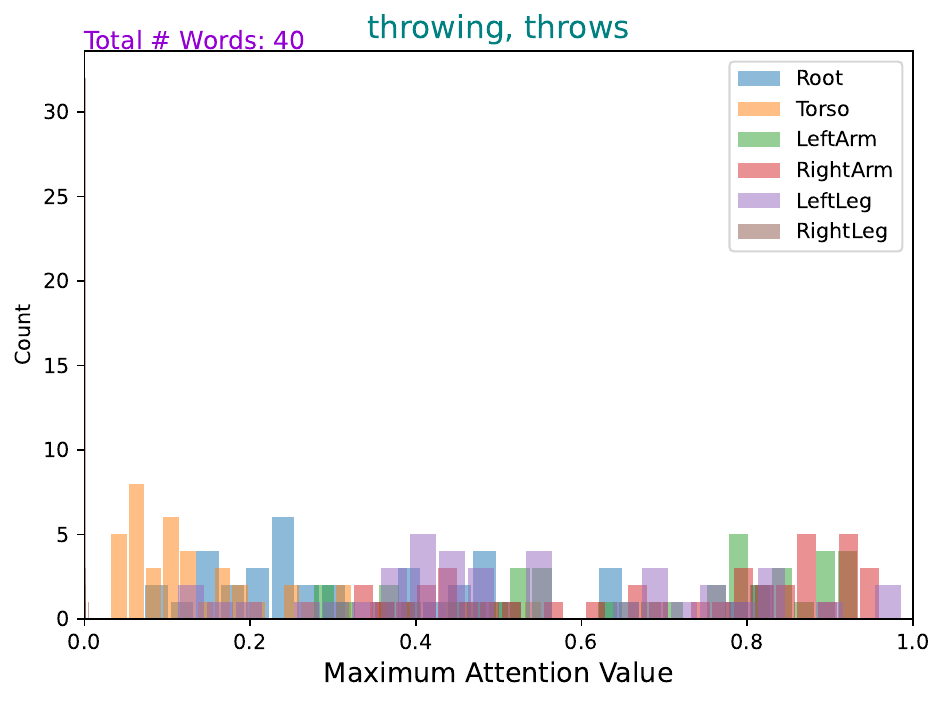}
        \caption{Without spatial supervision, attention incorrectly focused on legs rather than arms for "throw" motion in some cases (left leg).}
        \label{fig:w/o_spatt_h3D}
    \end{subfigure}\hfill
    \begin{subfigure}{0.48\columnwidth}
        \centering
        \includegraphics[width=\columnwidth,height=4cm]{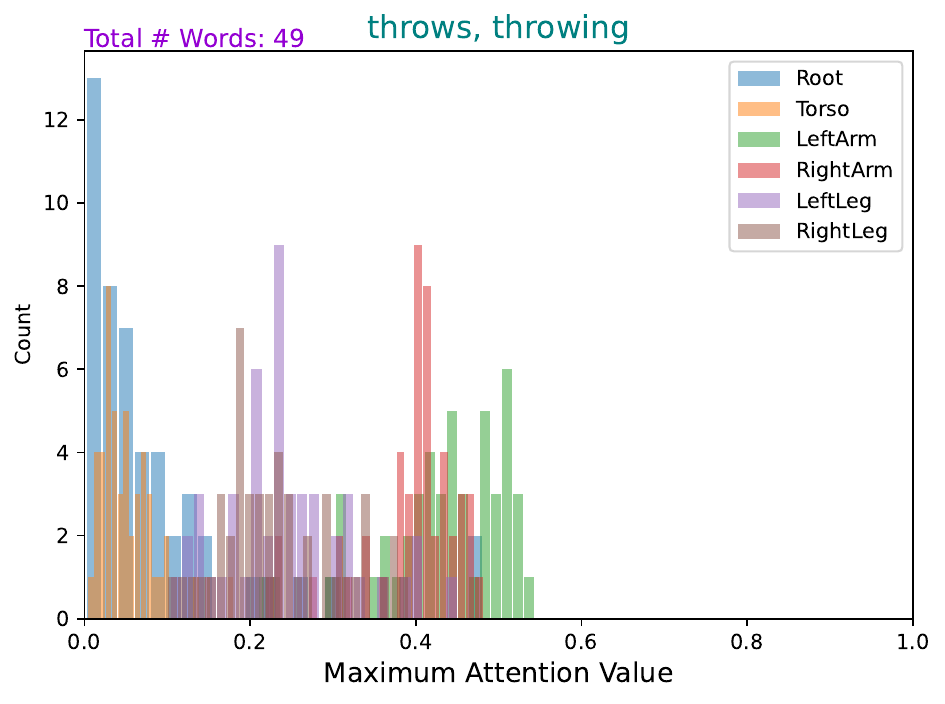}
        \caption{With spatial supervision, spatial attention is always maximal on relevant part, for this example on the arms.}
        \label{fig:w_spatt_h3D}
    \end{subfigure}
    \caption{Effect of spatial supervision on HumanML3D across the entire test set for a given motion word (e.g. \textit{throw}) (\# Refer to number of the given motion words).}
    \label{fig:w_vs_wo_spatt_h3D}
\end{figure}

\textbf{Body part identification.} We can illustrate the effectiveness of our architecture in learning a correct body part association through spatio-temporal attention by viewing the density distribution for maximum attention across time per each body part for some motion words 
as illustrated in Figures \ref{fig:w_vs_wo_spatt_h3D} and \ref{fig:spat_temp} (Diverse examples in Supp.\ref{supp:visualizations}).

\textbf{Action localization.} Another aspect that emerges from temporal Gaussian attention weights is action localization. The architecture shows ability to identify motion onset without temporal supervision. We can derive the action onset from spatio-temporal attention maps, as illustrated in \Cref{fig:TAG_figs} where we also show their actual onset time.

\FloatBarrier

\textbf{Transfer to adjacent tasks.} Similar tasks can benefit from the proposed methodologies. In the context of skeleton based action recognition and localization, our proposed motion encoder and skeleton partitioning could be used to build an interpretable model. In a continuous stream, action segmentation tasks could be also cast as sequence to sequence learning, thus attention weights could be used to infer the action start/end times as an unsupervised learning. If the action time is available, these annotations could serve to supervise the spread of temporal weights, further enhancing the accuracy of action localization and spatio-temporal attention maps. Given an image, for each visual word in the caption, our spatial supervision could be transformed into maximizing the attention weights on relevant objects. Finally, the interpretability could be evaluated using the proposed density function for adaptive attention and histograms for attention distribution on spatial locations in other captioning context.
 
\section{Conclusion}
\label{sec:conclusion}
We have introduced guided attention with adaptive gate for motion captioning. After evaluating the influence of different weighting schemes for the main loss terms, we have found that our approach leads to interpretable captioning while improving performance. Interpretability is very important to consider when designing an architecture, it's gives insights on model capability to perform a true reasoning. This ensures the ability of generalizing instead of memorizing. The proposed model addressed the two challenges, given an interpretable result with accurate semantic captions. The model and proposed methodology can be transposed to other captioning tasks, such as supervision of spatial attention weights in action recognition tasks.

\section*{Acknowldgements}
This work is supported by the Occitanie Region of France (Grant ALDOCT-001100 20007383) and the European Union's HORIZON Research and Innovation Programme (Grant 101120657, Project ENFIELD).

\bibliography{main}



\appendix
\renewcommand{\thesubsection}{\Alph{subsection}}
\section*{Supplementary}


This supplementary provides more details on the method implementation, and more visualization for global evaluation of interpretability. Furthermore, we discuss the effectiveness of architecture design. All following analysis are conducted on the test set.
For illustration, visual animations are included in the github repository\footnote{\url{https://github.com/rd20karim/M2T-Interpretable}}. The transparency level of the gold box represents the temporal attention variation for each predicted \textit{motion word} selected based on \textit{adaptive attention}. We note that grammatical errors mainly stem from the datasets themselves, which contain valid action descriptions but sometimes with incorrect language structure.

\section{Motivation}
\label{supp:motiv}

Our approach is \textbf{focusing on interpretability} while ameliorating motion captioning performance. This comes with additional challenging question on accurate methods for interpretability evaluations. To address this question, a first attempt is to draw multiple visualizations. However, for a global evaluation on test set, this become infeasible. To overcome this limitation, in addition, a simple solution, yet effective, is to display histogram and density distributions for attention weights across all test set instead of just sample wise visualizations.

The architectural design is primarily intended to be interpretable, allowing for the explanation of learned spatial, temporal, and adaptive attention weights. Designing an efficient architecture while maintaining interpretability can be very challenging, but has several advantages beyond focusing solely on increasing accuracy metrics. In addition to ensuring a reliable model, we can leverage the interpretability provided by attention mechanisms to extract other semantic motion information: action localization, body part and motion word identification. Let's recall the main novel contributions of our paper in this context:

\begin{itemize}
    \item Interpretable architecture design.
    \item Supervision of adaptive and spatial attention.
    \item Effective tools for global interpretability evaluation.
\end{itemize}

Consequently, regarding each contribution aspect, we will show the concrete effectiveness of associated theoretical formulations.

\section{Datasets}
\label{supp:dataset}
We use the two commonly used benchmarks KIT-MLD and Human ML3D with the following statistical details:

\begin{table}[ht]
\centering
\begin{tabular}{lcccc}
\toprule
\bf Subset    & \bf Number  & \bf Train & \bf Test & \bf Val. \\
\midrule
\multirow{2}{*}{KIT-ML-aug}& \ motions  & 4886  & 830 & 300      \\ 
                            &   samples & 10408  & 1660  & 636 \\
\midrule
\multirow{2}{*}{HML3D-aug}    & motions & 22068  &  4160  & 1386    \\ 
                              & samples & 66734 & 12558 & 4186 \\
                              \bottomrule
\end{tabular}
\caption{Data splits, for KIT and Human ML3D after augmentation (aug). }
\label{tab:kit_h3d_size}
\end{table}

\section{Ground-truth generation for supervision}
\label{supp:gt_generation}

\paragraph{Predefined dictionary.}

We manually define a dictionary based on representative words in the dataset describing different motion characteristics. Intentionally the dictionary doesn't cover all datasets actions with their synonyms, we want the model to be able to generalize to remaining unsupervised words for their spatial and gate attention. We will see later that the model effectively converges for this intended behavior.

\begin{table}[h]
    \centering
    \resizebox{\columnwidth}{!}{
    \begin{tabular}{c|c|c}
    \bf Category          & \bf Words                                                       & \bf Body part \\ [5pt] \hline
     \bf Trajectory  & circle, circuit, clockwise, anticlockwise, forward, backward     &  Root \\ [5pt]  \hline
    \bf \multirow{4}{*}{Local motion} & open, waves, wipe, throw, punch, pick, boxing, & \multirow{2}{*}{Arms}\\ [5pt] 
                                    &  clean, swipe, catch, handstand, draw &  \\ [5pt]  \cline{2-3}
                                     & kick, stomp, lift, kneel, squat, squad, stand, stumble, rotate & Legs \\ [5pt] \cline{2-3}
                                 & bend, bow & Torso \\ [5pt]  \hline
         \bf Connection words & is, the, of, his, her, its, on, their  & -\\ [5pt]  \hline
         \bf Subject & a, person, human, man & - \\
         \bottomrule
    \end{tabular}}
    \caption{Predefined dictionary for both datasets.}
    \label{tab:dict_words}
\end{table}

During training, the words in Table \ref{tab:dict_words}, and targets words, are stemmed to find correspondence for spatial weight supervision.

\paragraph{Spatial attention supervision.}

The ground truth spatial attention weights $\alpha_{ti}$ are generated based on the predefined dictionary 
and it's same for all frames, the temporal attention is the responsible for temporal filtering.

\paragraph{Adaptive attention supervision.} The ground truth $\beta_t$ is generated based on the Part Of Speech (POS) tagging.

\section{Hyperparameters selection}
\label{supp:hyperparam}
We run experiments for different values of $(\lambda_{spat},\lambda_{adapt})$. The quantitative results are reported in Table \ref{tab:sweep_hyp}.

\begin{table}[h]
    \centering
    \resizebox{\columnwidth}{!}{
    
    \begin{tabular}{lccccccc}
    \toprule
    \bf Dataset &  $\lambda_{\mbox{\bf spat}}$ & \bf $\lambda_{\mbox{\bf adapt}}$ &  \bf BLEU@1 & \bf BLEU@4 & \bf CIDEr & \bf ROUGE-L & \bf BERTScore \\ \midrule
        \multirow{6}{*}{KIT-ML} & 0 & 0 & 57.3 & 23.6 & 109.9 & 57.8 & 41.1 \\ 
         & 0 & 3 & 56.3 & 22.5 & 108.4 & 56.5 & 39.8 \\
         & 1 & 3 & 57.6 & 23.5 & 102.6 & 57.2 & 40.1 \\
         & 2 & 3 & \textbf{58.4} & \textbf{24.4} & \textbf{112.1} & \textbf{58.3} & \textbf{41.2} \\ 
         & 3 & 5 & 57.6 & 23.7 & 105.7 & 57.5 & 40.9 \\
         & 5 & 5 & 56.5 & 22.0 & 99.4 & 56.8 & 39.9 \\ \hline \hline
        \multirow{8}{*}{HML3D} & 0 & 0 & 69.3 & 24.0 & 58.8 & 54.8 & 38.7 \\ 
                                 & 0 & 3 & \textbf{69.9} & \textbf{25.0} & 61.6 & \textbf{55.3} & \textbf{40.3} \\
                                 & 0.1 & 3 & 69.5 & 23.8 & 58.7 & 55.0 & 38.9 \\ 
                                 & 0.25 & 3 & 68.7 & 23.8 & 59.7 & 54.7 & 39.3 \\
                                 & 0.5 & 3 & 68.8 & 23.8 & 60.0 & 55.0 & 38.6 \\ 
                                 & 1 & 3 & 68.7 & 23.7 & 58.2 & 54.6 & 39.0 \\ 
                                 & 2 & 3 & 69.2 & 24.4 & \textbf{61.7} & 55.0 & 40.3 \\ 
                                 & 3 & 3 & 68.3 & 23.2 & 56.5 & 54.5 & 37.1 \\ 
                                 \bottomrule
         \end{tabular} }
\caption{Spat+adapt supervision impact w.r.t each corresponding weights.}    
\label{tab:sweep_hyp}
\end{table}

\FloatBarrier
\section{Architecture compounds effectiveness}
\label{supp:abla_archi}

We aim in the following visualizations to demonstrate the global effectiveness of architecture design of each compound :

\begin{itemize}
    \item Functionality of gating mechanism.
    \item Impact of Part based motion encoding.
    \item Spatio-temporal attention blocks.
\end{itemize}

\paragraph{Gating mechanism.}

The gate variable $\beta$ allows the model to use or not the motion information given the word time step. To visualize this internal process of switching between motion and language, we display predictions for the best model on KIT-ML (results on HumanML3D were shown in the paper). As we see in the following Table, the context vector ($\beta=1$) is successfully  used for all motion characteristics: \textit{action}, \textit{speed}, \textit{body parts}, \textit{trajectory}, \textit{direction}…. Particularly, we note that the end token \verb|<eos>| is also motion related, as outputting this word depends on the end of the relevant human motion range.

\begin{figure}[t]
    \centering
    \includegraphics[scale=0.25]{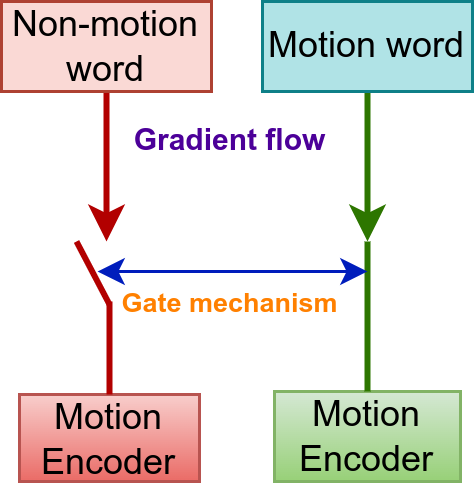}
    \caption{Illustration of our gating mechanism during training. This mechanism prevent the decoder from attending to motion for non-motion word. Consequently the motion encoder is prevented from receiving important gradients updates for non motion words.}
    \label{fig:enter-label}
\end{figure}

\FloatBarrier

\paragraph{Spatial+adapt attention supervision [KIT-ML].}
We show comparison of Spatio-temporal attention maps and text generated between the case of supervision and w/o supervision:

\begin{figure}[h]
    \centering
    \begin{subfigure}{\textwidth}
    \includegraphics[width=\textwidth]{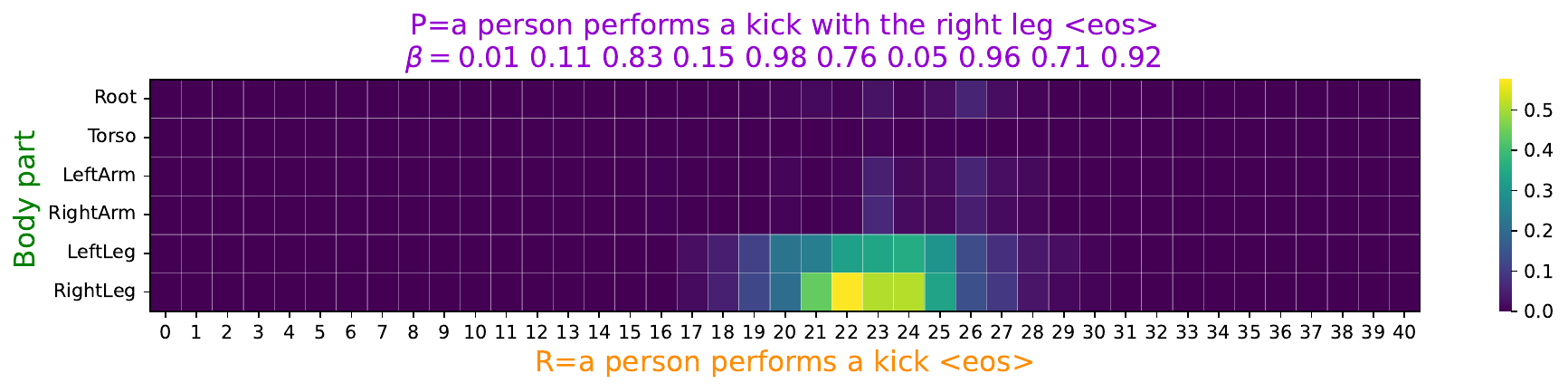}
    \caption{With  supervision KIT-(2,3) (action range [19,28]/right kick).}
    \label{fig:w_superv_att}\vfill     \vspace{1em}

    \begin{subfigure}{\textwidth}
        \includegraphics[width = \textwidth]{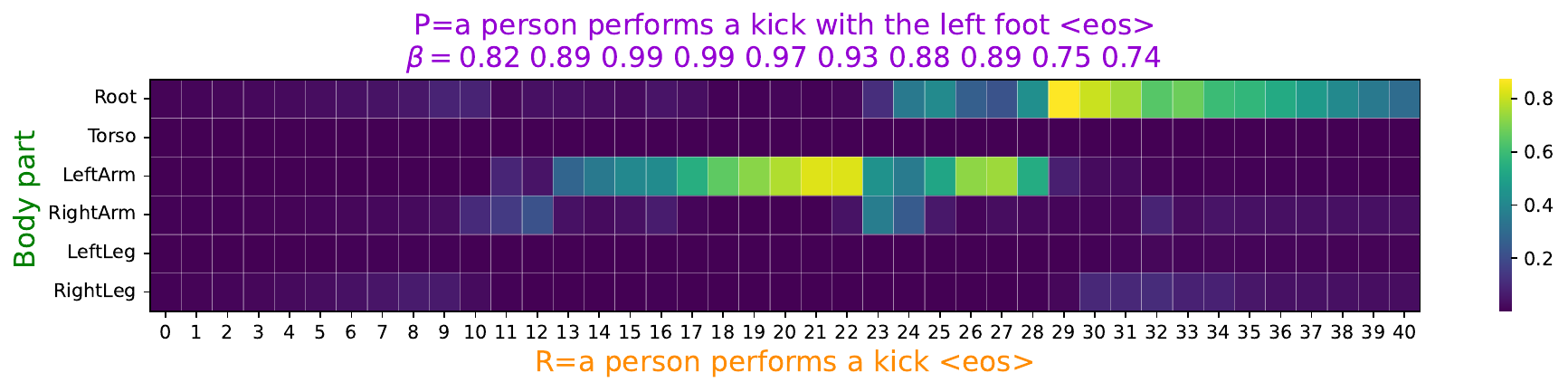}
        \caption{Without supervision KIT-(0,0) (action range [19,27]/right kick).}
        \label{fig:w/o_superv_att}
    \end{subfigure} 

    \end{subfigure} 
\end{figure}

As we see in the case of supervision (Fig.\ref{fig:w_superv_att}) the part were correctly identified and perfectly localized in the range $[20,26]$ with corresponding manually identified range $[19,28]$ and small $\beta$ values are associated with non-motion words. Without supervision (Fig.\ref{fig:w/o_superv_att}), the model focuses on irrelevant part and consequently the range of action was not precisely localized. Additionally the $\beta$ values are high for all kind of words.

\begin{figure}[h]
    \centering
    \begin{subfigure}{\textwidth}
        \includegraphics[width = \textwidth]{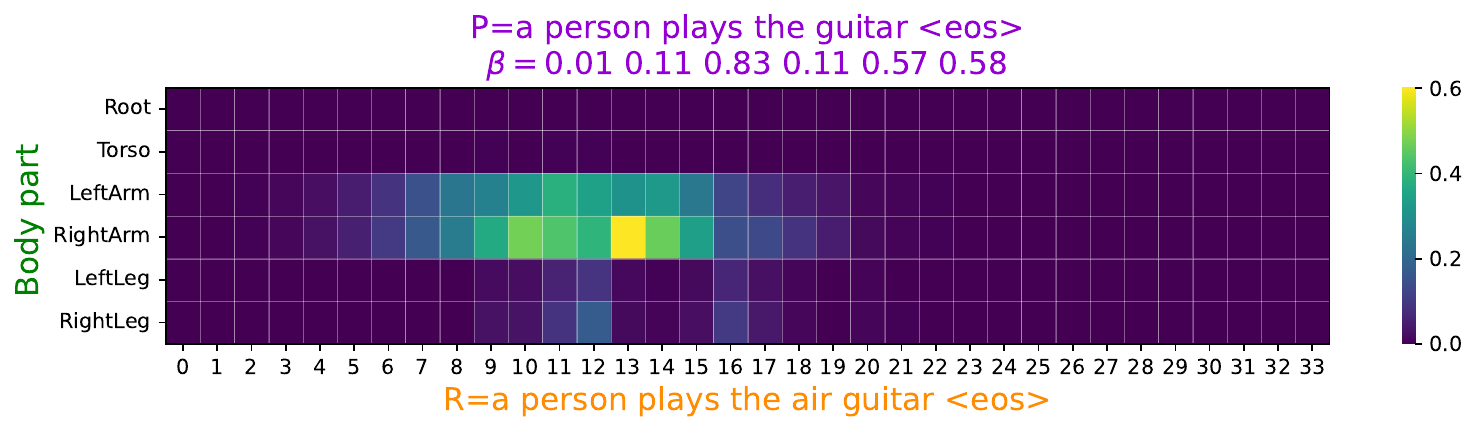}
        \caption{Play (action range [10,20]).}
        \label{fig:subfig1}
    \end{subfigure} \vfill  \vspace{1em}

    \begin{subfigure}{\textwidth}
        \includegraphics[width=\textwidth]{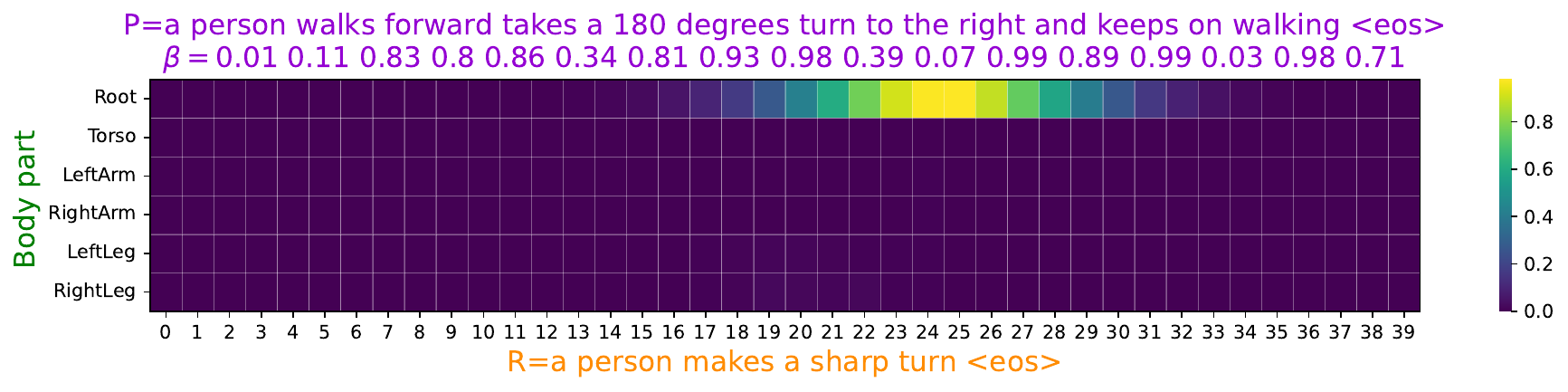}
        \caption{Turn (action range [22,27]).}
        \label{fig:subfig2}
    \end{subfigure} \vfill \vspace{1em}

    \begin{subfigure}{\textwidth}
    \includegraphics[width=\textwidth]{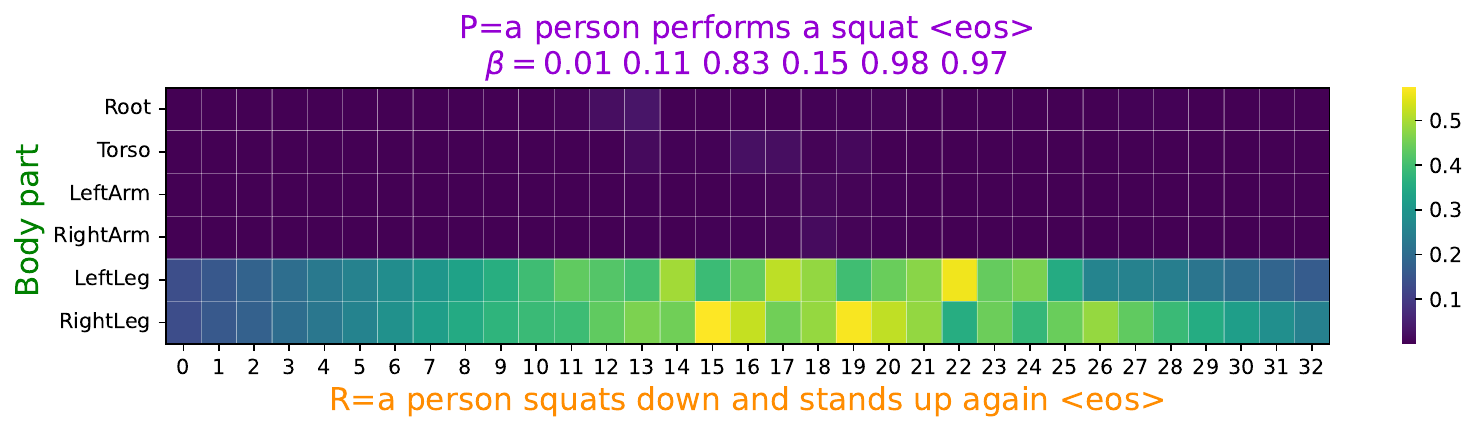}
    \caption{Squat (action range [10,28]).}
    \label{fig:subfig2}
\end{subfigure} 
\caption{Spatio-temporal attention for different motion words on KIT-ML.}
\label{fig:spatemp_kit}
\end{figure}

We visualize more samples (Fig.\ref{fig:spatemp_kit}) with Spatial+adapt supervision. Temporal range is mentioned for comparison, even if action localization wasn't the main focus in captioning task, the model was able to learn implicitly a temporal location through the temporal Gaussian attention mechanism. 

\paragraph{Trajectory and global motion.}
 
\begin{figure}[h]
    \centering
    \begin{subfigure}[b]{0.48\textwidth}
        \includegraphics[width=\textwidth]{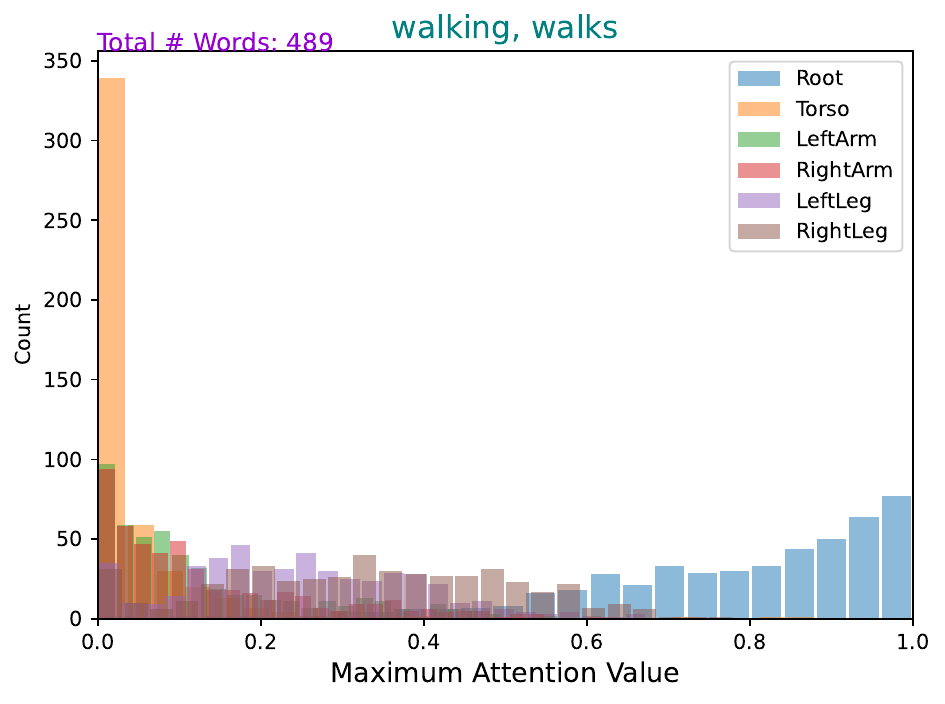}
        \caption{Walk}
        \label{fig:subfig1}
    \end{subfigure}
    \hfill
    \begin{subfigure}[b]{0.48\textwidth}
        \includegraphics[width=\textwidth]{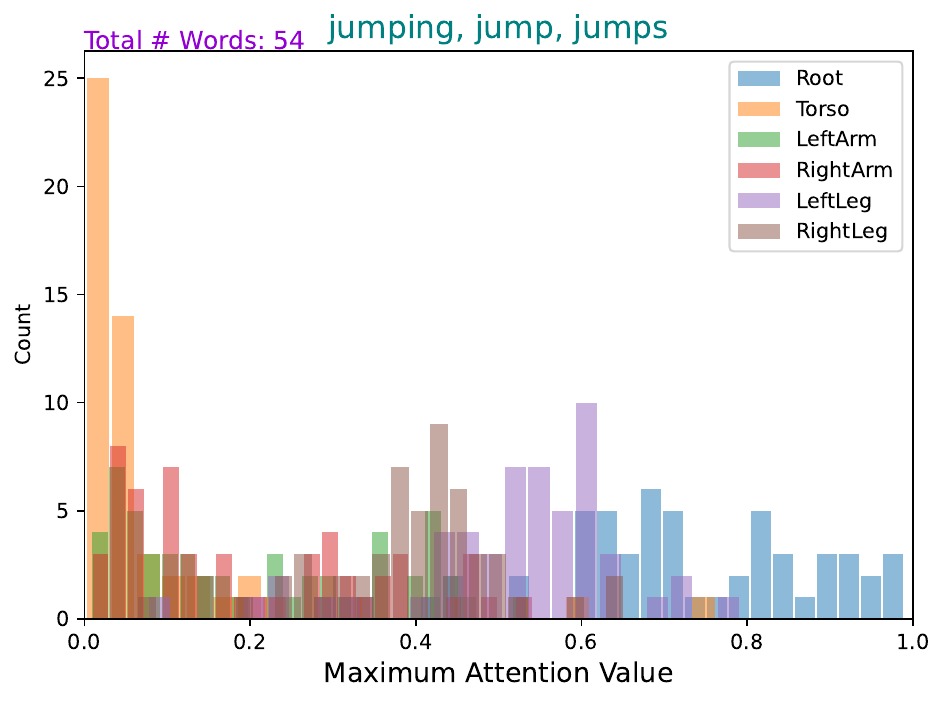}
        \caption{Jump}
        \label{fig:subfig2}
    \end{subfigure}
    \caption{[KIT-(2,3)]: Body part distribution (spat+adapt).}
    \label{fig:kit_hist}
\end{figure}

 The attention was supervised only for words describing trajectory, but the model generalize successfully to motion words highly depending on global trajectory. This result on maximum attention distributed toward the \textit{Root} body part, as we see in Figure \ref{fig:kit_hist}.

\FloatBarrier

\section{Part based encoding \& spatio-temporal attention}
\label{supp:visualizations}
As mentioned in the paper, our architecture design could be sufficient in learning a correct spatial attention maps using larger dataset with rich semantic descriptions. For demonstration, we will use the model with no spatial supervision, to show that part based encoding and spatio-temporal can work solely and correctly together for focusing on relevant body parts w.r.t to the associated generated motion word.
To this purpose, we propose to display the histogram distribution of temporal maximum attention weights for each body part over all test set and given a different motion words. This allows for an effective global evaluation of interpretability over all test set.

\paragraph{Histograms.} In the following, we display the body parts histogram distribution across the test set for different motion words on the model with \textbf{\textit{no spatial supervision}} as demonstration for the effectiveness in finding relevant parts to focus on using our interpretable architecture design that includes part-based encoding along with spatio-temporal attention.  This is only in the case of the larger dataset HumanML3D.  The KIT-ML small dataset still requires spatial supervision to help the architecture focusing on relevant part, as the vocabulary and its size are limited. As demonstrated in all following Figures, depending on the motion word, arms-based/legs-based actions, and particularly some motions with an emphasis on Torso body part.

\begin{figure}
    \centering     
    \begin{minipage}[b]{0.48\textwidth}
        \includegraphics[width=\textwidth]{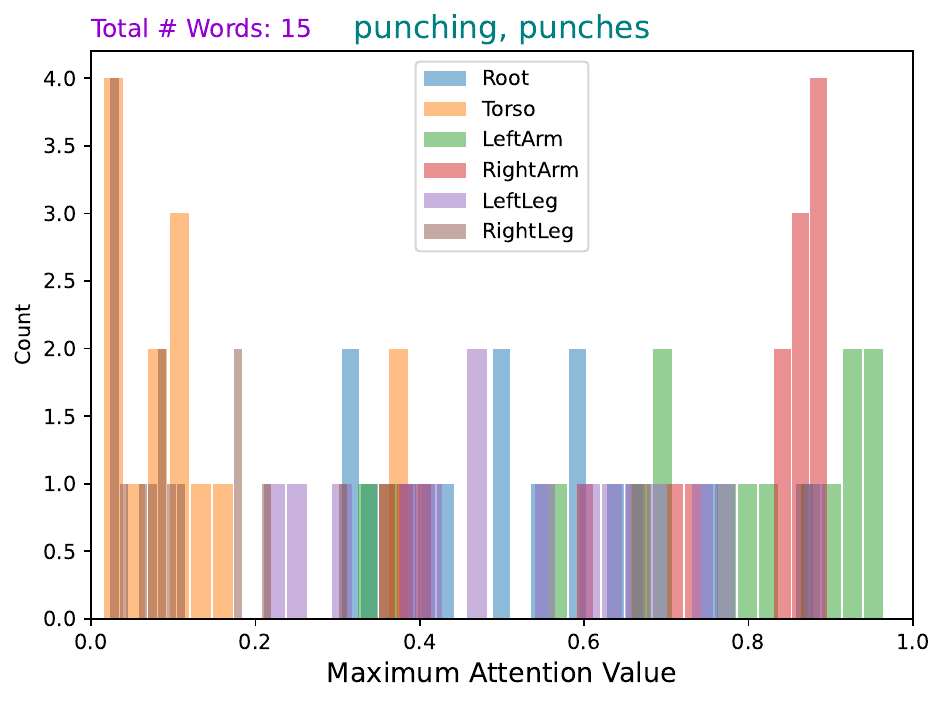}
        \caption*{Punch}
        \label{fig:subfig1}
    \end{minipage}
    \hfill
    \begin{minipage}[b]{0.48\textwidth}
        \includegraphics[width=\textwidth]{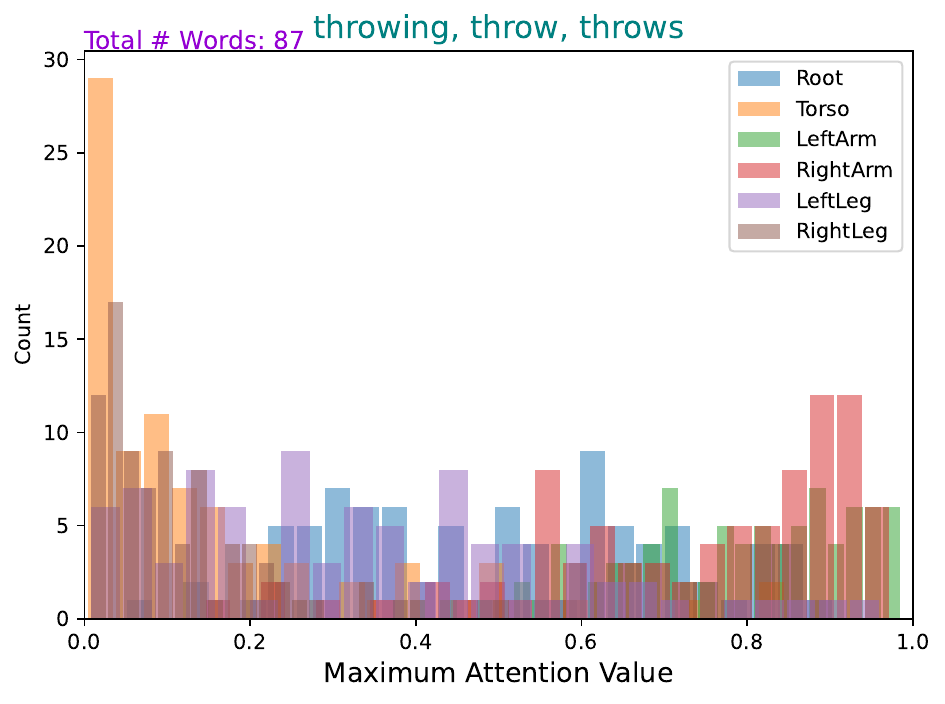}
        \caption*{Throw}
        \label{fig:subfig2}
    \end{minipage}
    
    \vspace{1em}
    
    \begin{minipage}[b]{0.48\textwidth}
        \includegraphics[width=\textwidth]{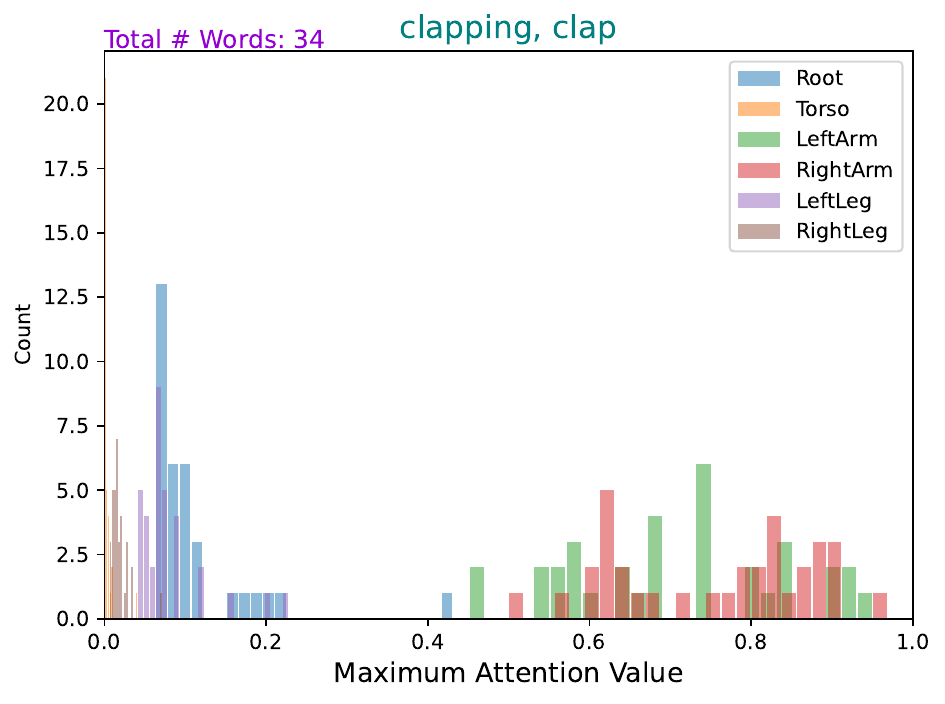}
        \caption*{Clap}
        \label{fig:subfig3}
    \end{minipage}
    \hfill
    \begin{minipage}[b]{0.48\textwidth}
        \includegraphics[width=\textwidth]{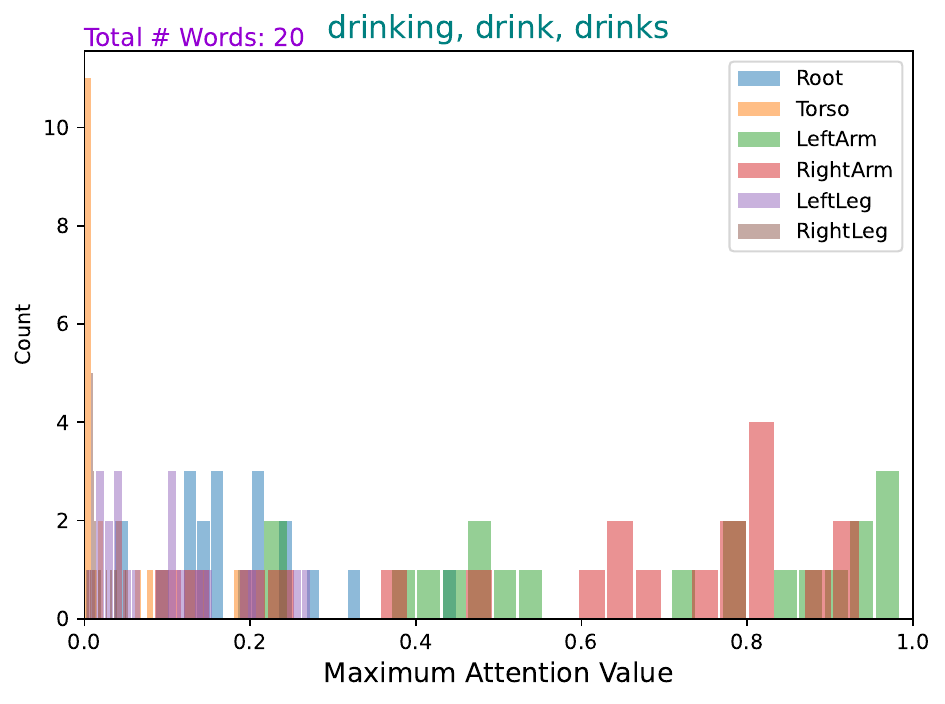}
        \caption*{Drink}
        \label{fig:subfig4}
    \end{minipage}
    
    \vspace{1em}
    
    \begin{minipage}[b]{0.48\textwidth}
        \includegraphics[width=\textwidth]{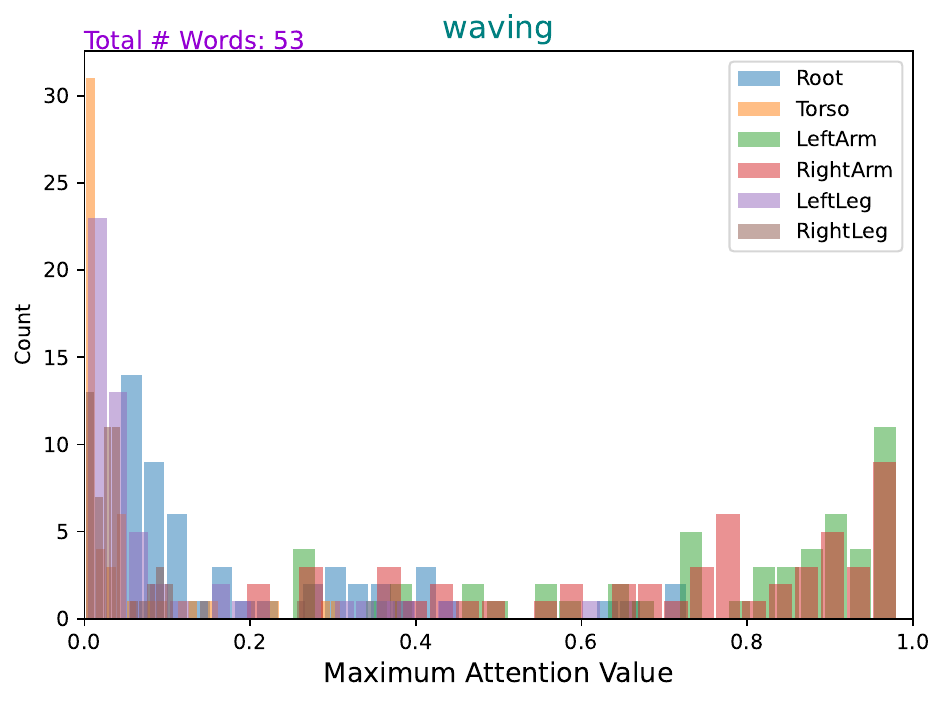}
        \caption*{Wave}
        \label{fig:subfig5}
    \end{minipage}
    \hfill
    \begin{minipage}[b]{0.48\textwidth}
        \includegraphics[width=\textwidth]{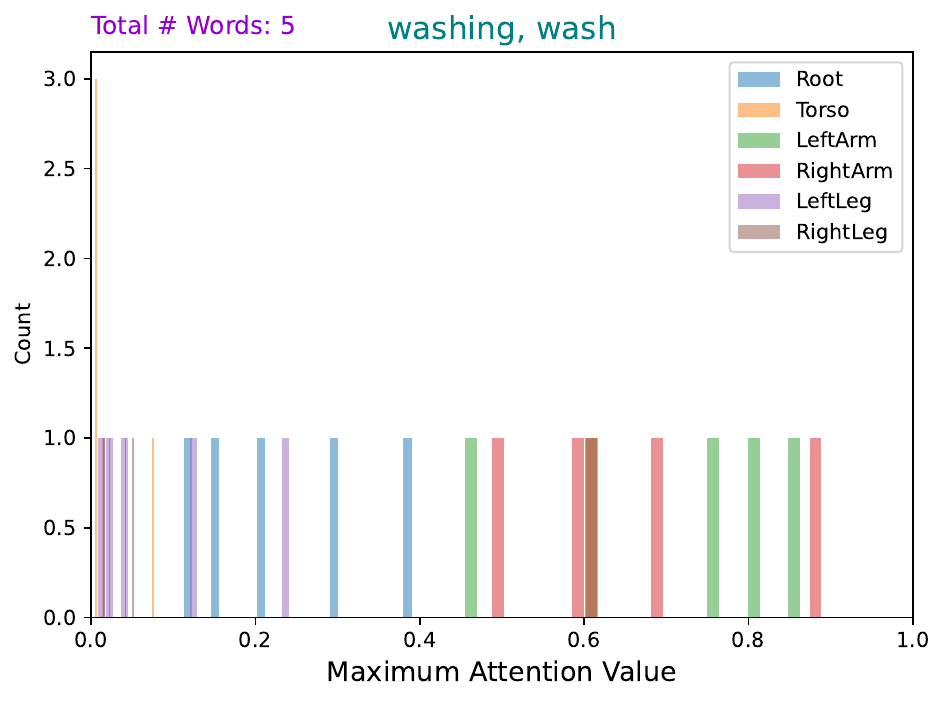}
        \caption*{Wash}
        \label{fig:subfig6}
    \end{minipage}
    
    \caption{Histogram generated on the HML3D with the config (0,3).}
    \label{fig:mainfigure}
\end{figure}

\begin{figure}[h]
    \centering

    \begin{minipage}[b]{0.48\textwidth}
        \includegraphics[width=\textwidth]{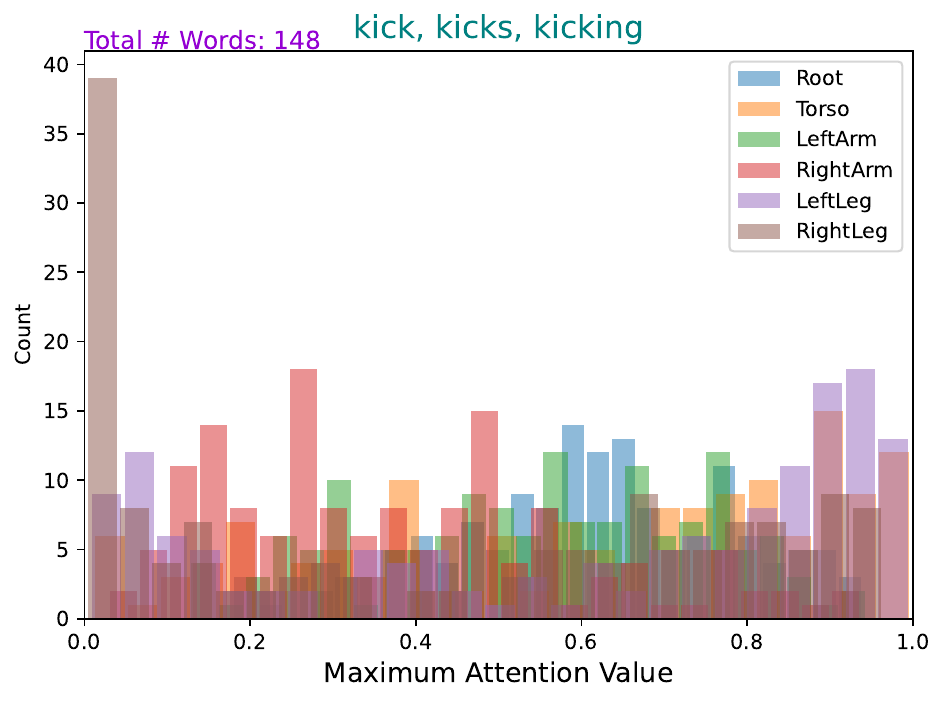}
        \caption*{Kick}
        \label{fig:subfig1}
    \end{minipage}
    \hfill
    \begin{minipage}[b]{0.48\textwidth}
        \includegraphics[width=\textwidth]{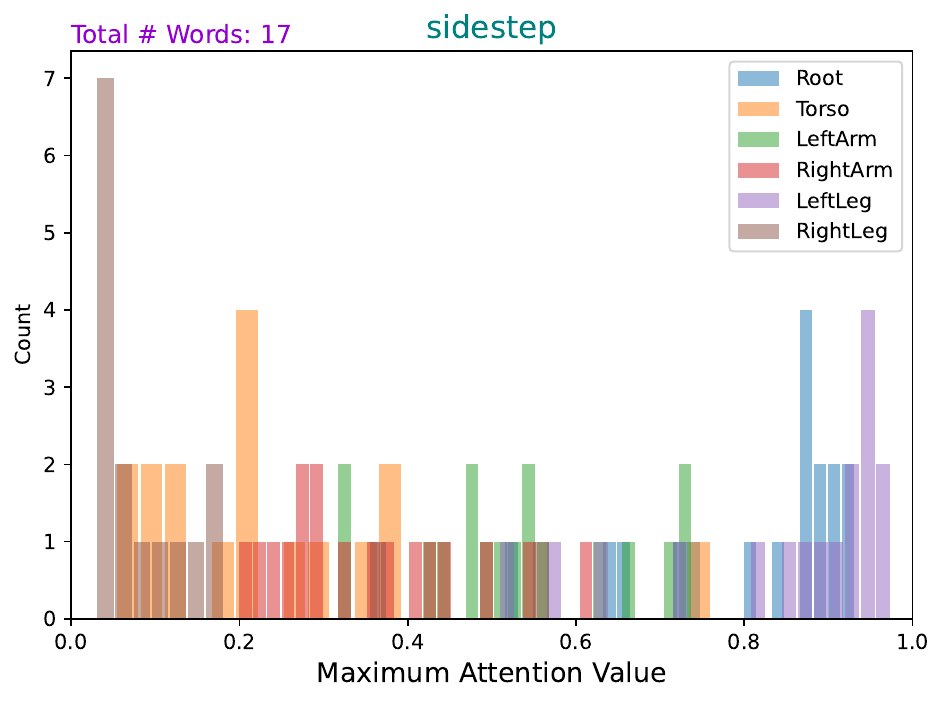}
        \caption*{Sidestep}
        \label{fig:subfig2}
    \end{minipage}

    \vspace{1em}

    \begin{minipage}[b]{0.48\textwidth}
        \includegraphics[width=\textwidth]{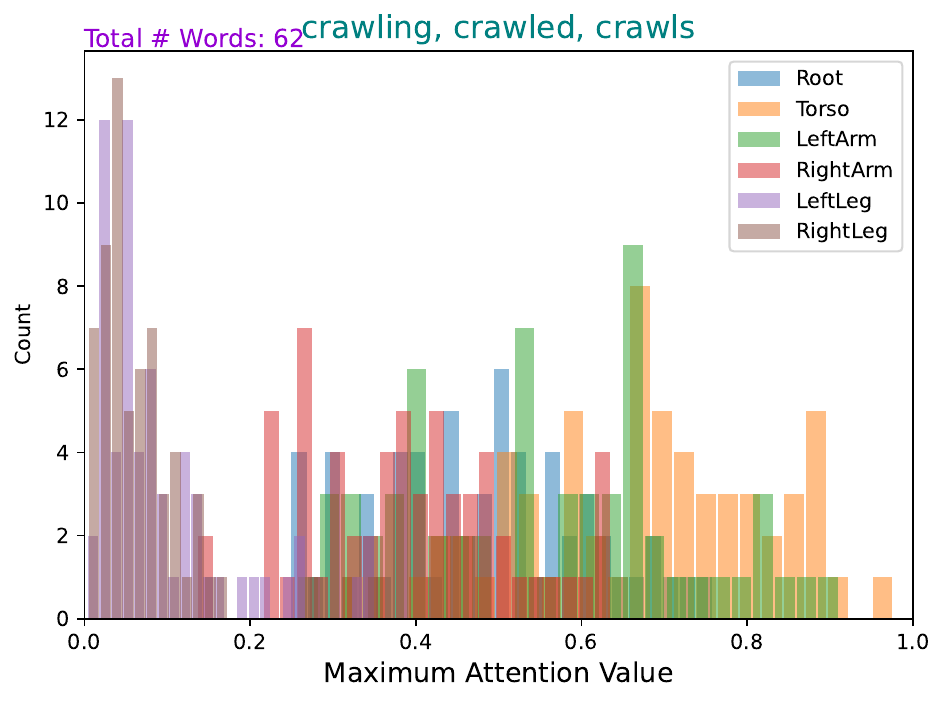}
        \caption*{Crawl}
        \label{fig:subfig3}
    \end{minipage}
    \hfill
    \begin{minipage}[b]{0.48\textwidth}
        \includegraphics[width=\textwidth]{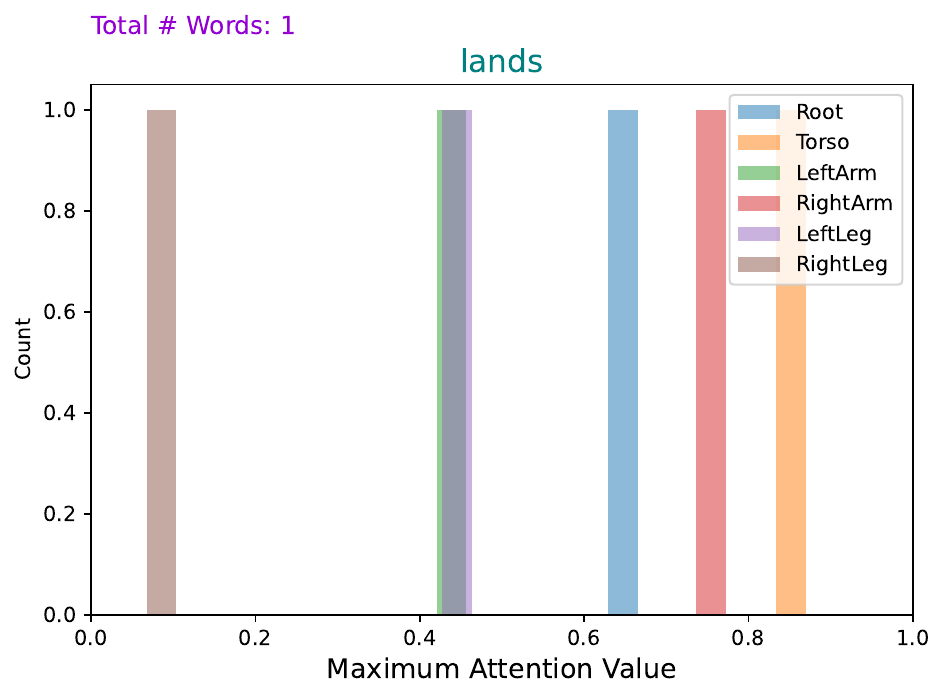}
        \caption*{Land}
        \label{fig:subfig4}
    \end{minipage}

    \vspace{1em}

    \begin{minipage}[b]{0.48\textwidth}
        \includegraphics[width=\textwidth]{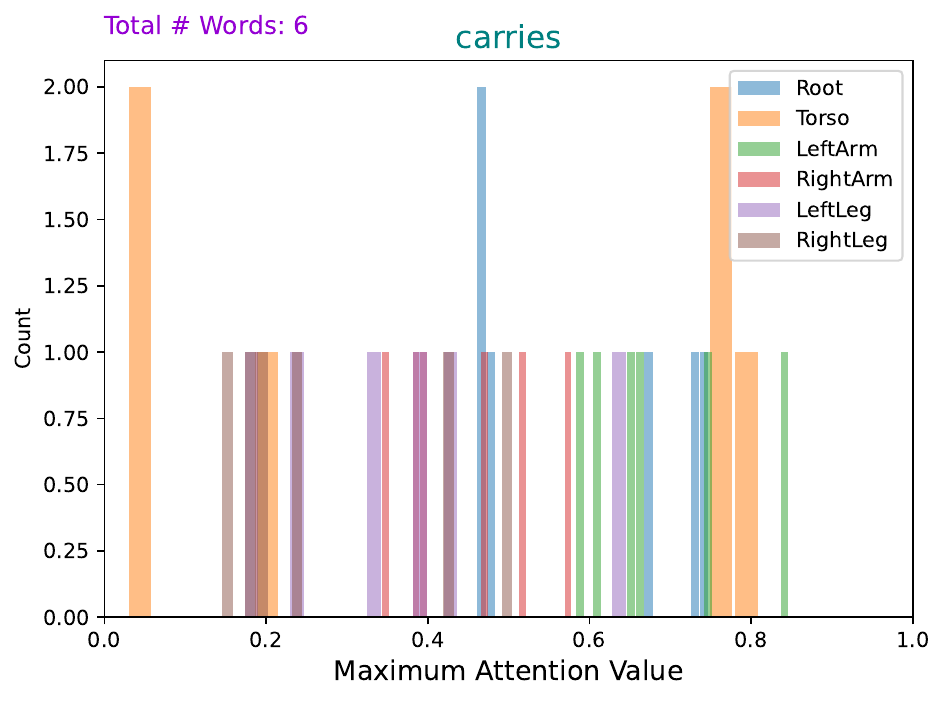}
        \caption*{Carries}
        \label{fig:subfig6}
    \end{minipage}
    \hfill
    \begin{minipage}[b]{0.48\textwidth}
        \includegraphics[width=\textwidth]{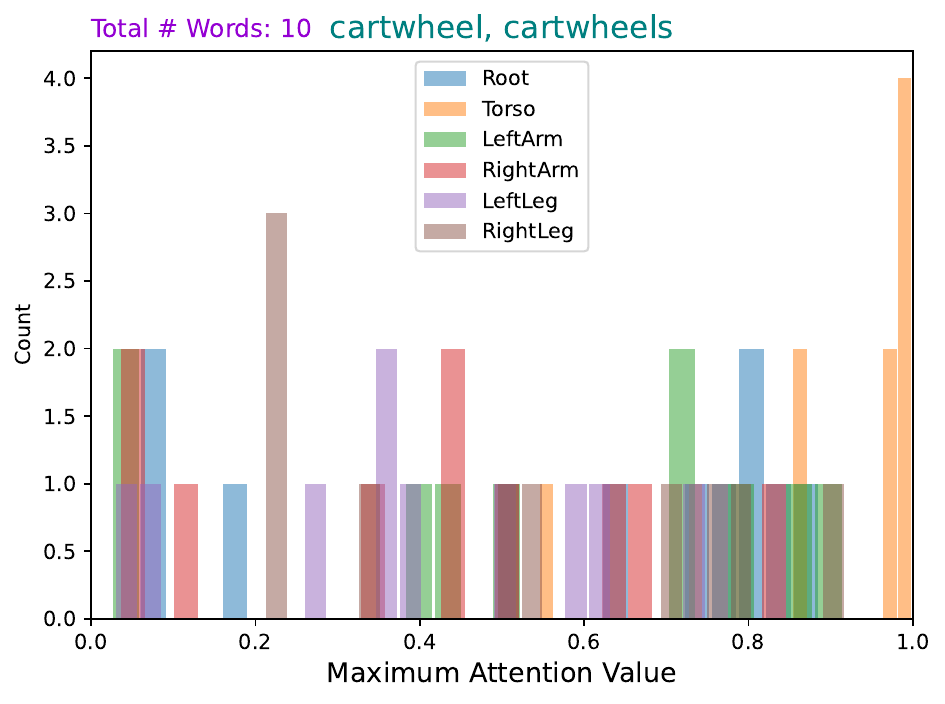}
        \caption*{Cartwheel}
        \label{fig:subfig8}
    \end{minipage}

    \caption{Histogram generated on HML3D with the config (0,3).}
    \label{fig:mainfigure}
\end{figure}

\FloatBarrier

\begin{figure}[p]
    \centering
    \begin{minipage}[b]{0.48\textwidth}
        \includegraphics[width=\textwidth]{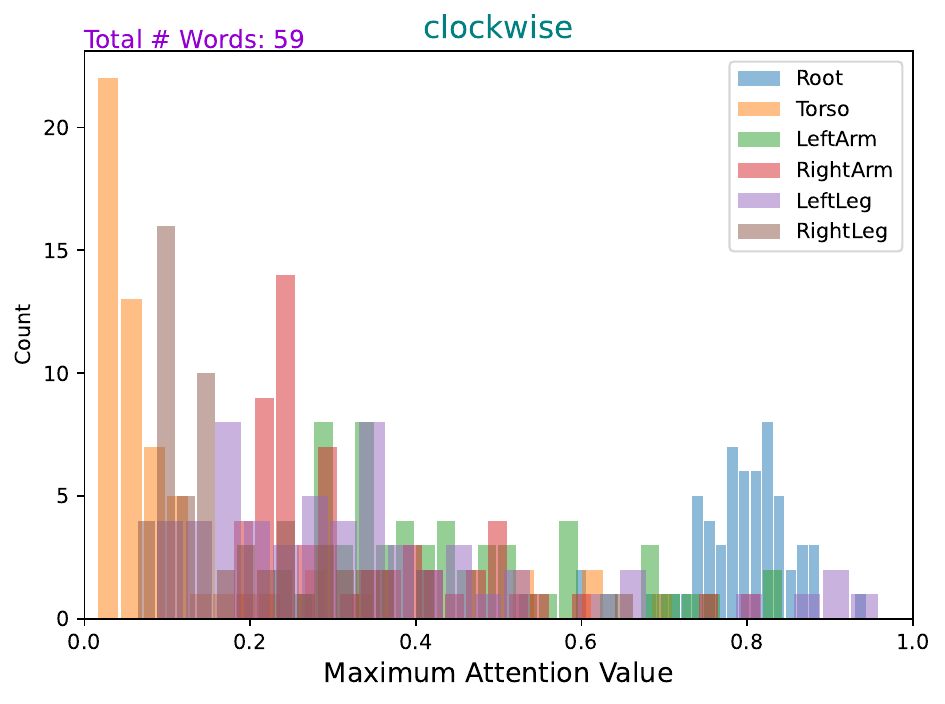}
        \caption*{Clockwise}
        \label{fig:subfig5}
    \end{minipage}
    \hfill
    \begin{minipage}[b]{0.48\textwidth}
        \includegraphics[width=\textwidth]{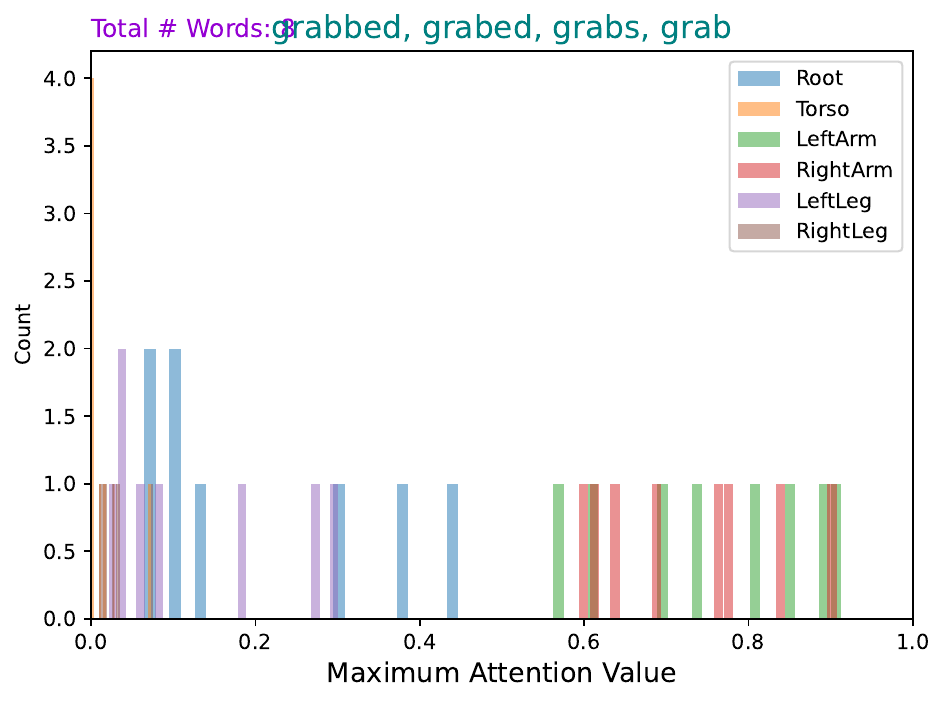}
        \caption*{Grab}
        \label{fig:subfig7}
    \end{minipage}

    \vspace{1em}

    \begin{minipage}[b]{0.48\textwidth}
        \includegraphics[width=\textwidth]{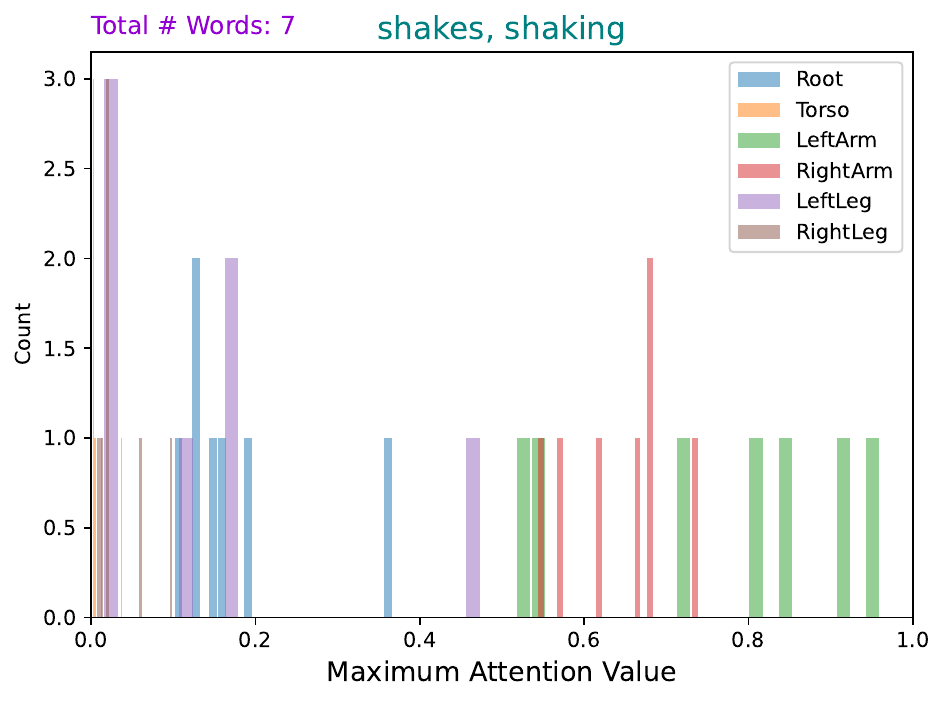}
        \caption*{Shake}
        \label{fig:subfig9}
    \end{minipage}
    \hfill
    \begin{minipage}[b]{0.48\textwidth}
        \includegraphics[width=\textwidth]{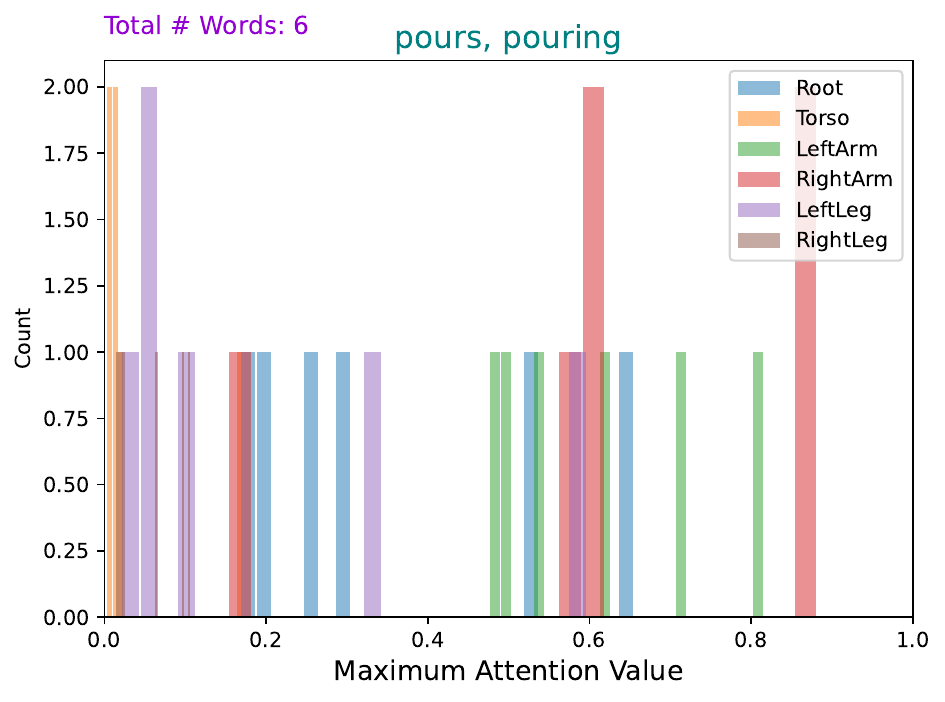}
        \caption*{Pour}
        \label{fig:subfig10}
    \end{minipage}

    \caption{Histogram generated on HML3D with the config (0,3).}
    \label{fig:mainfigure2}
\end{figure}

\FloatBarrier
\newpage
\paragraph{Spatio-temporal attention maps.}

In this part, we display attention maps for some interesting words for HML3D (0-3) /adapt. 
In the case of the model without spatial supervision, we have found that the model performs a correct attention focus. When an action is performed using right leg/arm, the model focuses correctly on the corresponding parts. Moreover, for actions performed with both arms/legs, the model focus on both parts. For all cases, body part words (left/right/both) are always accurately identified into the generated text. These observations are common across different representative samples (from different actions).

\begin{figure}[h]
    \centering
    \begin{minipage}{\textwidth}
        \centering
        \includegraphics[width=0.9\textwidth]{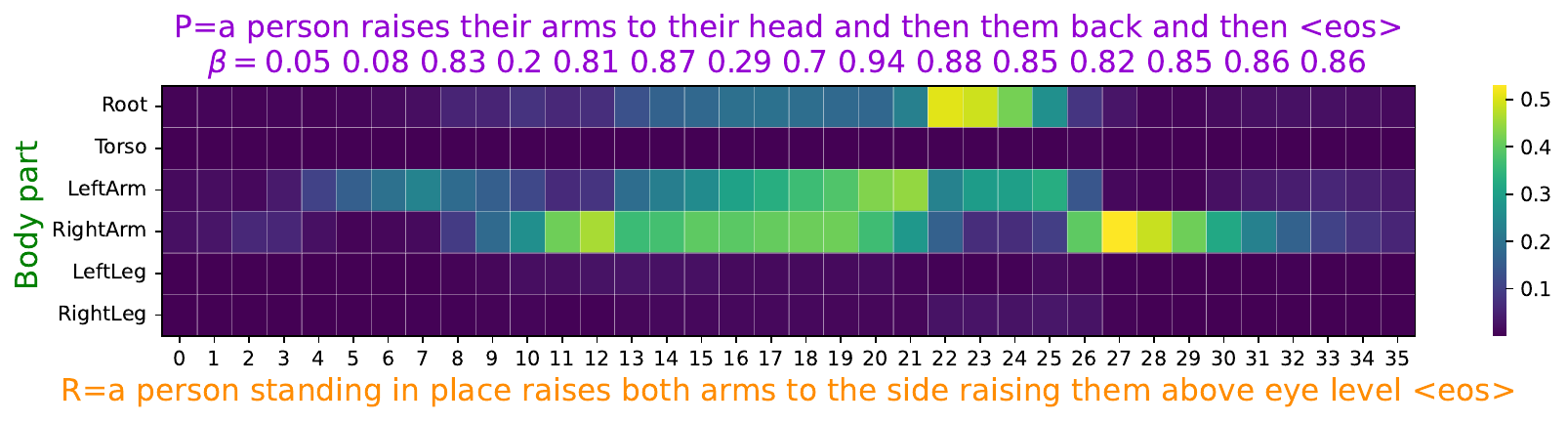}
        \caption*{Raises.}
        \label{fig:subfig2}
    \end{minipage}
    
    \vspace{1em}
    
    \begin{minipage}{\textwidth}
        \centering
        \includegraphics[width = 0.9\textwidth]{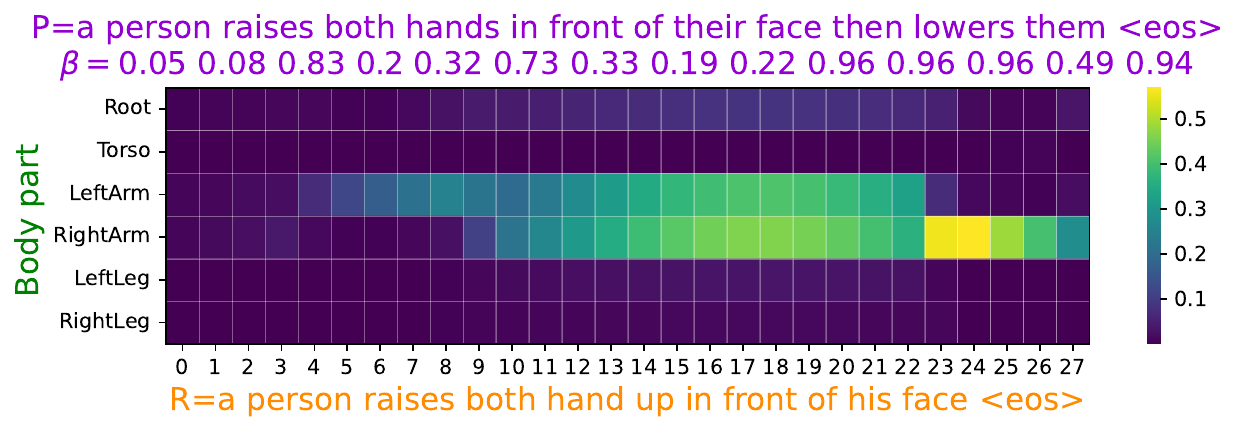}
        \caption*{Lowers.}
        \label{fig:subfig1}
    \end{minipage} 
\end{figure}

\newpage
\FloatBarrier
\vspace{-9cm}
\begin{figure}[h]
    \centering
    \begin{minipage}{\textwidth}
        \centering
        \includegraphics[width = 0.75\textwidth]{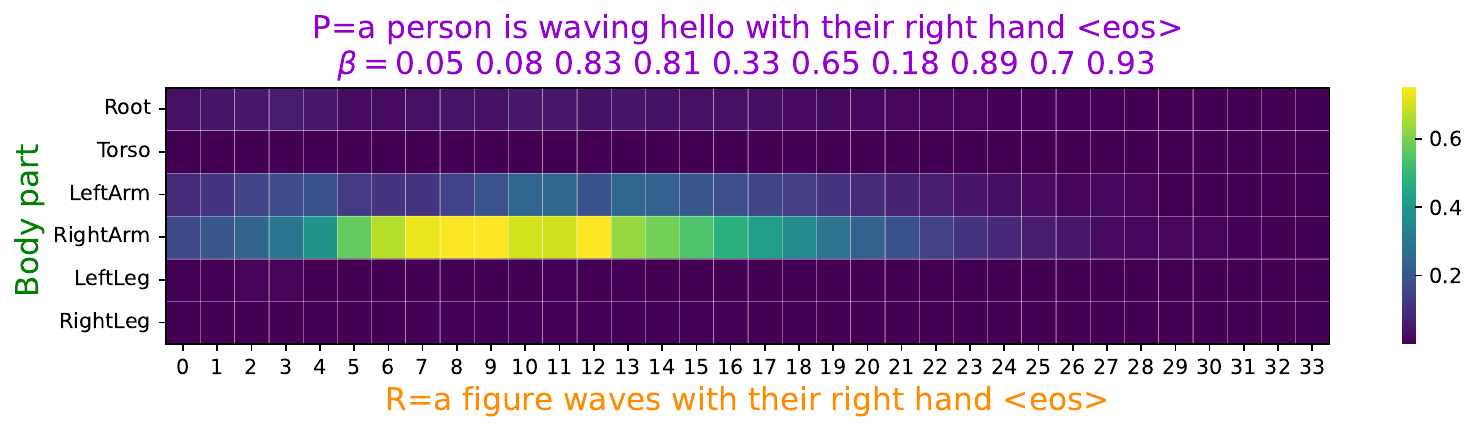}
        \caption*{Waving.}
        \label{fig:subfig3}
    \end{minipage}
    
    \vspace{1em}
    
    \begin{minipage}{\textwidth}
        \centering
        \includegraphics[width=0.75\textwidth]{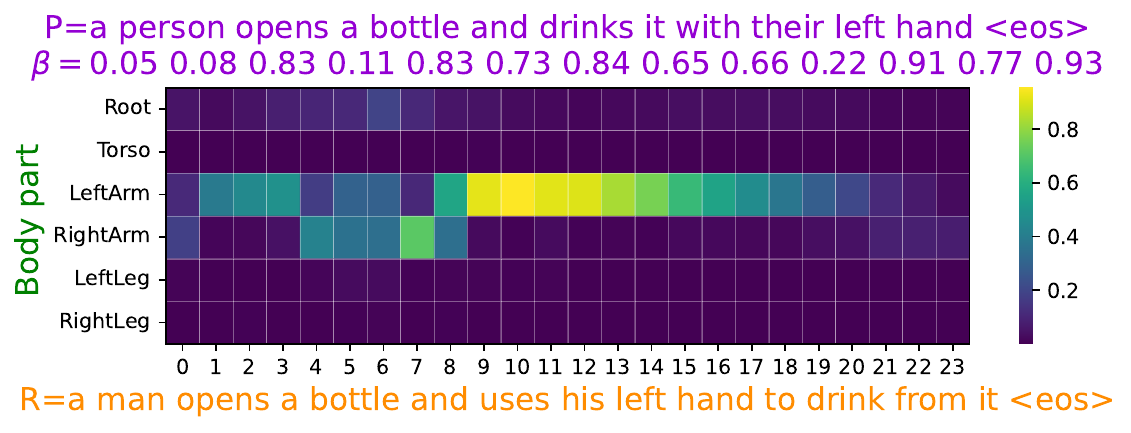}
        \caption*{Opens.}
        \label{fig:subfig4}
    \end{minipage}
    
    \vspace{1em}
    
    \begin{minipage}{\textwidth}
        \centering
        \includegraphics[width = 0.75\textwidth]{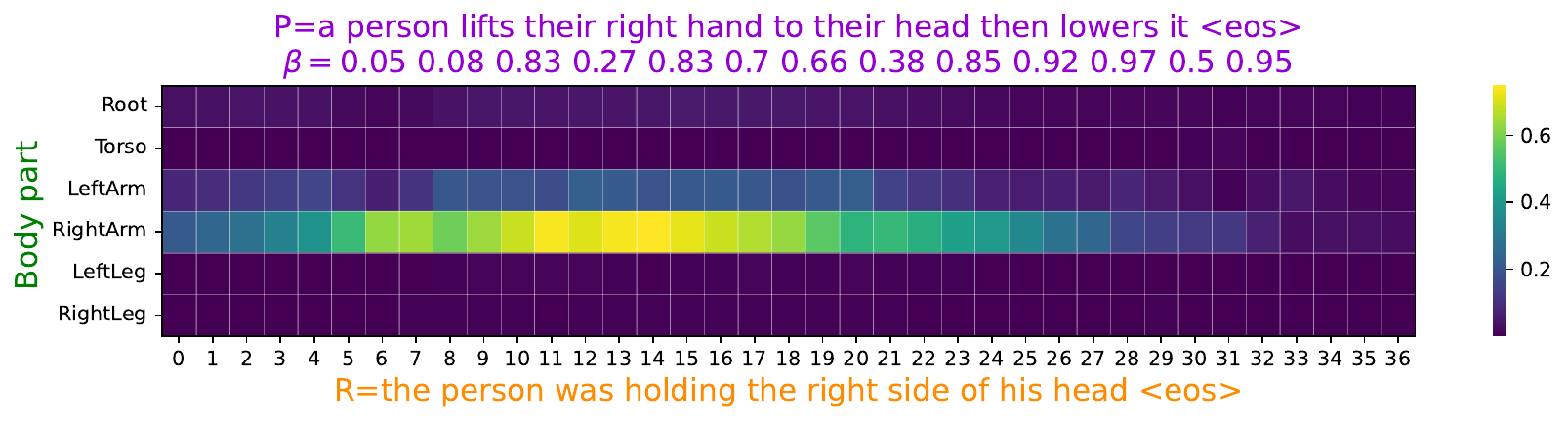}
        \caption*{Lifts.}
        \label{fig:subfig5}
    \end{minipage}
    
    \vspace{1em}
    
    \begin{minipage}{\textwidth}
        \centering
        \includegraphics[width = 0.75\textwidth]{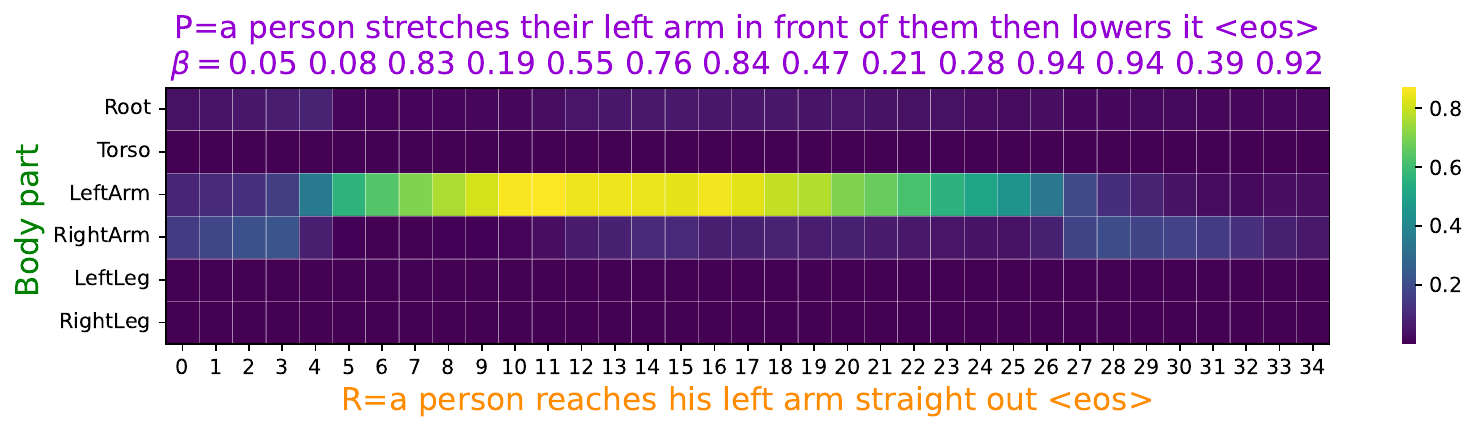}
        \caption*{Stretches.}
        \label{fig:subfig6}
    \end{minipage}
\end{figure}



\end{document}